\documentclass[10pt,twocolumn,letterpaper]{article}

\PassOptionsToPackage{table}{xcolor}
 \usepackage[pagenumbers]{cvpr} 

\makeatletter
\@namedef{ver@everyshi.sty}{2001/05/15 v3.00 everyshi package (MS)}
\makeatother
\usepackage{makecell}
\usepackage{multirow}
\usepackage{pifont}
\usepackage{bbding}
\usepackage[misc]{ifsym}
\usepackage{bm}
\usepackage{ulem}

\newcommand\blfootnote[1]{%
  \begingroup
  \renewcommand\thefootnote{}\footnote{#1}%
  \addtocounter{footnote}{-1}%
  \endgroup
}

\makeatletter
\@ifundefined{safe@activestrue}{\newif\ifsafe@active}{}
\makeatother
\usepackage{tikz}
\usepackage{float}
\usepackage{mdframed}

\newcommand{\cmark}{\ding{51}}
\newcommand{\xmark}{{\color{lightgray}\ding{55}}}
\definecolor{camelGreen}{rgb}{0.13, 0.55, 0.13}
\definecolor{babyblue}{rgb}{0.54, 0.81, 0.94}
\definecolor{iccvblue}{rgb}{0.0, 0.4, 0.8}  

\usepackage{rotating}
\usepackage{array}  

\newcolumntype{H}{>{\setbox0=\hbox\bgroup}c<{\egroup}@{}} 

\definecolor{cvprblue}{rgb}{0.21,0.49,0.74}
\usepackage[pagebackref,breaklinks,colorlinks,allcolors=cvprblue]{hyperref}

\def\paperID{}
\def\confName{CVPR}
\def\confYear{2026}

\title{From Detection to Association: Learning Discriminative Object Embeddings for Multi-Object Tracking}

\author{
{Yuqing Shao$^{1,2*}$, Yuchen Yang$^{3,2*}$, Rui Yu$^{1*}$, Weilong Li$^{4}$, Xu Guo$^{3,2}$,}\\
{Huaicheng Yan$^{1\dagger}$, Wei Wang$^{2\dagger}$, Xiao Sun$^{2}$}\\
\\
$^{1}$East China University of Science and Technology\ \
$^{2}$Shanghai AI Laboratory\ \
$^{3}$Fudan University \\
$^{4}$Sun Yat-sen University\\
}

\begin{document}
\maketitle
\blfootnote{Work done during internship at Shanghai AI Laboratory.}
\blfootnote{$^{*}$Equal contribution. $^{\dagger}$Corresponding authors.}

\begin{abstract}
  End-to-end multi-object tracking (MOT) methods have recently achieved remarkable progress by unifying detection and association within a single framework.
  Despite their strong detection performance, these methods suffer from relatively low association accuracy.
  Through detailed analysis, we observe that object embeddings produced by the shared DETR architecture display excessively high inter-object similarity, as it emphasizes only category-level discrimination within single frames.
  In contrast, tracking requires instance-level distinction across frames with spatial and temporal continuity, for which current end-to-end approaches insufficiently optimize object embeddings.
  To address this, we introduce FDTA (\textbf{F}rom \textbf{D}etection \textbf{t}o \textbf{A}ssociation), an explicit feature refinement framework that enhances object discriminativeness across three complementary perspectives.
  Specifically, we introduce a Spatial Adapter (SA) to integrate depth-aware cues for spatial continuity, a Temporal Adapter (TA) to aggregate historical information for temporal dependencies, and an Identity Adapter (IA) to leverage quality-aware contrastive learning for instance-level separability.
  Extensive experiments demonstrate that FDTA achieves state-of-the-art performance on multiple challenging MOT benchmarks, including DanceTrack, SportsMOT, and BFT, highlighting the effectiveness of our proposed discriminative embedding enhancement strategy.
  The code is available at \url{https://github.com/Spongebobbbbbbbb/FDTA}.
\end{abstract}
\section{Introduction}
\label{sec:intro}
\begin{figure}[!t]
    \centering
    \includegraphics[width=\linewidth]{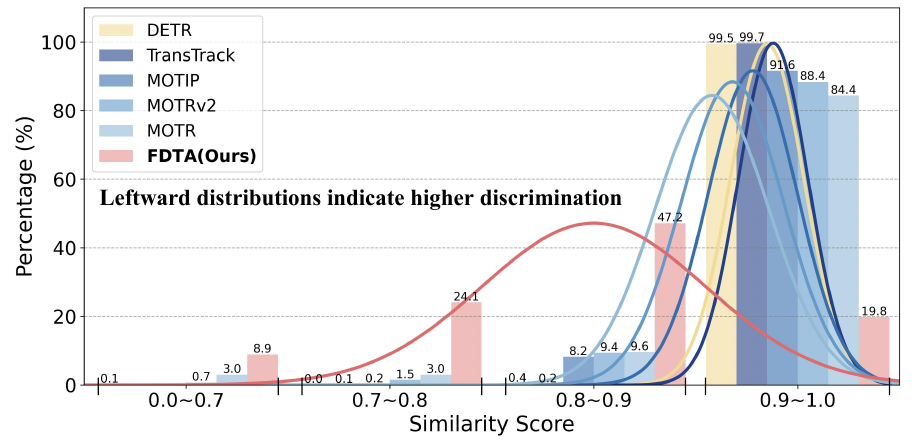}
    \caption{Embedding similarity analysis on DanceTrack. For each frame, we compute pairwise similarities between objects and select the top-3 highest values among all pairs, then construct their distribution across all frames. FDTA produces object embeddings with significantly lower similarity compared to existing methods.}
    \label{fig:intro}
    \vspace{-2mm}
\end{figure}

Multi-object tracking (MOT) jointly detects multiple objects in video sequences and associates them across frames with consistent identities. This technique is critical for a wide range of applications, including autonomous driving~\cite{hu2023planning,yu2020bdd100k}, action analysis~\cite{cioppa2022soccernet,caba2015activitynet,yang2025sga}, and robotics~\cite{fu2024mobile,ma2025learning}. 
Recent \textit{end-to-end} MOT methods~\cite{gao2023memotr,zeng2022motr,gao2024multiple,yan2023bridging,galoaa2024more,segu2024samba} leverage DETR~\cite{carion2020end,zhu2020deformable,liu2022dab} to generate object embeddings and perform detection and association within an integrated framework, achieving impressive results across multiple challenging benchmarks.

However, end-to-end methods suffer from low association accuracy (AssA$\sim$60\%), despite achieving high detection performance (DetA$>80\%$).
To investigate the cause, we examine the object embeddings produced by the shared DETR architecture.
As shown in \cref{fig:intro}, object embeddings from existing methods exhibit severe inter-object similarity, with over 80\% of inter-object similarity scores exceeding 0.9. Such high similarity blurs the distinction between different objects, further impairing the association during identification.
Notably, the similarity distribution of object embeddings in current end-to-end methods closely mirrors that of the original DETR, which is pretrained purely for detection.
The observation suggests an insufficient optimization of discriminative object embeddings for association.

\begin{figure}[t]
    \centering
    \includegraphics[width=\linewidth]{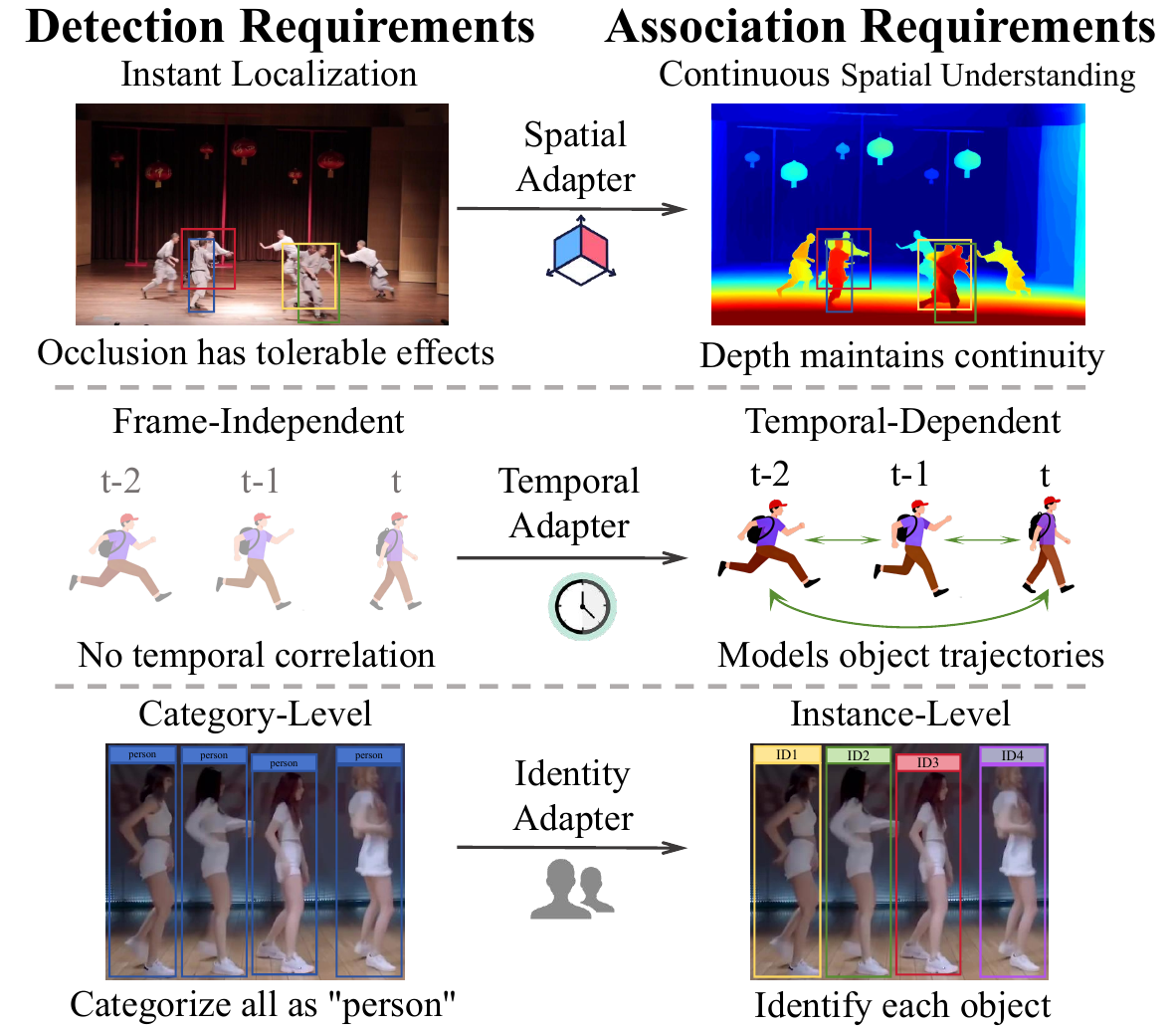}
    \caption{Illustration of the requirements of detection and association. It differs across spatial, temporal, and identity perspectives.}
    \label{fig:end2end}
    \vspace{-2mm}
\end{figure}

To address this issue, we aim to explicitly refine the object embeddings to enhance their discriminativeness. For effective and targeted refinement, we first clarify the fundamental differences in requirements between detection and association.
As shown in \cref{fig:end2end}, these requirements diverge in three key aspects: (1) \textbf{Spatial}: detection relies on instant localization, whereas association demands continuous spatial understanding; (2) \textbf{Temporal}: detection treats frames independently, while association requires global context understanding; (3) \textbf{Identity}: detection emphasizes category-level discrimination (\textit{person} vs. \textit{car}), while association necessitates instance-level differentiation (\textit{person \#1} vs. \textit{person \#2}).
Due to these disparities, object embeddings learned purely from the detection objective tend to be less discriminative, underscoring the need for additional optimization to make them suitable for association.

Based on the above analysis, we propose \textbf{FDTA} (\textbf{F}rom \textbf{D}etection \textbf{t}o \textbf{A}ssociation), an explicit feature refinement framework that learns discriminative embeddings via three complementary modules, each targeting one key aspect.

\noindent{(1) \textbf{S}patial \textbf{A}dapter (\textbf{SA})}. Depth information complements appearance features extracted for detection, mitigating occlusions by introducing spatial continuity. In SA, we construct a parallel feature extraction branch and introduce depth estimation as an auxiliary task, supervised by pseudo labels from a foundation model~\cite{chen2025video}. The depth-aware features are then fused with the original object embeddings to enhance spatial discriminativeness.

\noindent{(2) \textbf{T}emporal \textbf{A}dapter (\textbf{TA}).} 
In TA, temporal dependencies are modeled at the trajectory level, enabling each object embedding to aggregate information across its entire history. Specifically, we design an attention mask tailored for association, which accounts for both causality and occlusion, ensuring reliable temporal interactions between frames.
As a result, TA enhances the discriminativeness of object embeddings through temporal modeling.

\noindent{(3) \textbf{I}dentity \textbf{A}dapter (\textbf{IA}).} 
To achieve instance-level identification beyond category-level discrimination, IA introduces quality-aware contrastive learning on object embeddings, which naturally aligns with the objective of distinguishing instances.
Specifically, it leverages IoU to assess sample quality for guiding identity learning.
This contrastive learning encourages embeddings of the same identity to be pulled together while pushing apart those of different identities.

By refinement via three targeted modules, FDTA effectively learns object embeddings suited for tracking, exhibiting stronger discriminative power.
As shown in ~\cref{fig:intro}, FDTA significantly reduces inter-object similarity.
Consequently, FDTA achieves state-of-the-art performance across multiple benchmarks~\cite{sun2020dancetrack, cui2023sportsmot, zheng2024nettrack}, especially on key tracking metrics HOTA and IDF1.

Our contributions are summarized as follows:
\begin{itemize}
    \item We identify a common limitation in existing end-to-end MOT methods: the produced object embeddings from DETR exhibit excessively high inter-object similarity, revealing insufficient optimization for tracking.
    \item We propose FDTA, an explicit feature refinement framework that effectively enhances object discriminativeness through three targeted adapters in spatial, temporal, and identity perspectives.
    \item Extensive experiments demonstrate FDTA achieves state-of-the-art performance across various benchmarks, thereby confirming the effectiveness of discriminative object embeddings in improving tracking performance.
\end{itemize}

\section{Related Work}
\label{sec:related_work}
Following prior works~\cite{xu2021transcenter,sun2020transtrack,gao2023memotr,yan2023bridging,galoaa2024more}, we categorize multi-object tracking methods into two main paradigms: \textit{tracking-by-detection} and \textit{end-to-end}.
\begin{figure*}[t]
    \centering
    \includegraphics[width=\linewidth]{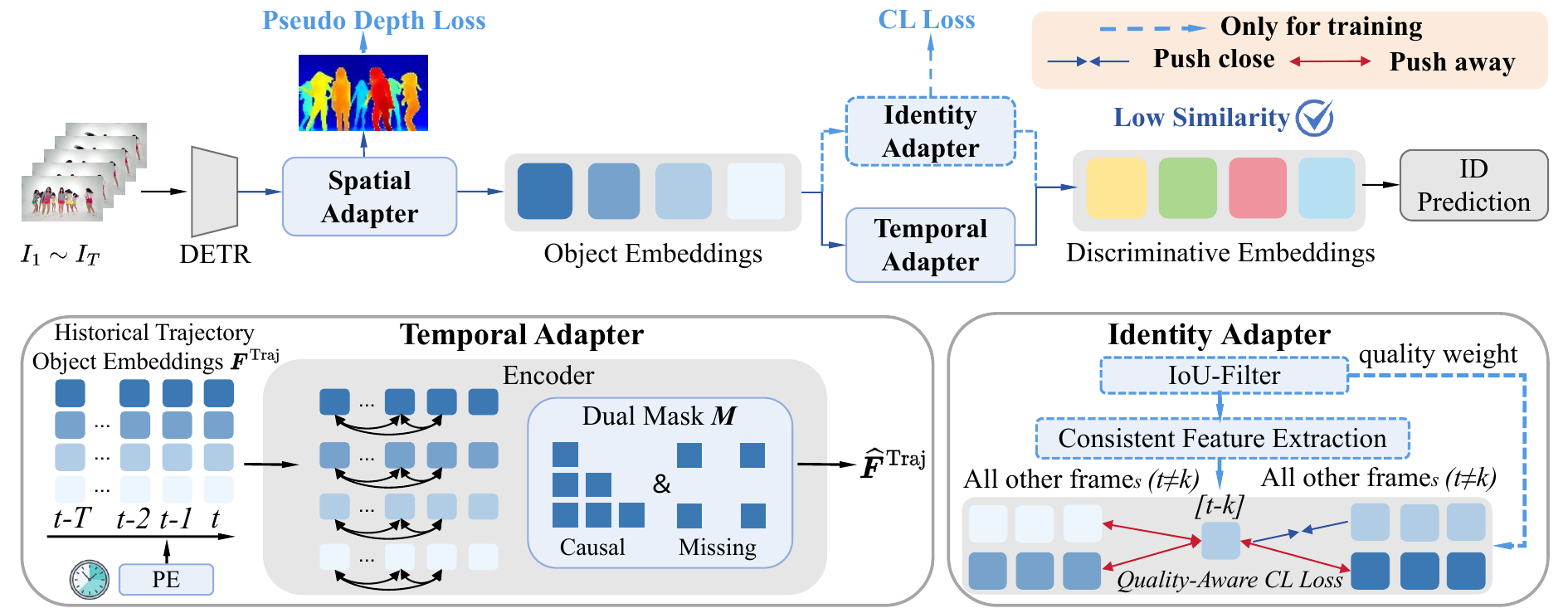}
    \caption{Overview of the FDTA framework.
    DETR produces object embeddings from input frames.
    The object embeddings are then refined by three explicit adapters for discriminativeness: Spatial Adapter (SA)
    integrates 3D geometric cues via depth learning; Temporal 
    Adapter (TA) captures temporal dependencies via trajectory modeling; 
    Identity Adapter (IA) promotes instance-level identification 
    via contrastive learning.
    Finally, an ID Prediction module performs the object association based on the enhanced embeddings.}
    \label{fig:framework}
  \end{figure*}

\noindent\textbf{Tracking-by-Detection.} This paradigm decouples detection and association into two sequential stages.
Pretrained detectors first generate bounding boxes, which are then linked across 
frames by association modules~\cite{bewley2016simple,zhang2022bytetrack,cao2022observation,chu2023transmot,zhou2022global}. Early methods primarily focus on association algorithms. SORT~\cite{bewley2016simple} introduces Kalman filtering, and subsequent works enhance robustness by integrating Re-ID features for appearance matching~\cite{wojke2017simple,zhang2021fairmot}, compensating for camera motion~\cite{cao2022observation}, recovering low-confidence detections~\cite{zhang2022bytetrack}, and refining local matching strategies~\cite{shim2025focusing}.
More recent approaches, such as TransMOT~\cite{chu2023transmot} and GTR~\cite{zhou2022global}, employ spatiotemporal transformers for learning-based association.
Despite these advances, the decoupled design still inevitably suffers from information loss and error propagation between stages.

\noindent
\textbf{End-to-End.} To overcome the information loss in the decoupled tracking-by-detection paradigm, several methods~\cite{zeng2022motr,meinhardt2022trackformer,gao2023memotr,zhang2023motrv2,galoaa2024more,gao2024multiple} employ DETR~\cite{carion2020end,zhu2020deformable,liu2022dab} to jointly optimize detection and association by generating unified object embeddings, forming an end-to-end paradigm.
MOTR~\cite{zeng2022motr} uses RNN-like sequential processing to auto-regressively propagate track queries across frames. 
Building on this, MOTRv2~\cite{zhang2023motrv2} improves query initialization using YOLOX~\cite{ge2021yolox}, while MeMOT~\cite{cai2022memot} and MeMOTR~\cite{gao2023memotr} incorporate memory banks to better aggregate trajectory history. MOTE~\cite{galoaa2024more} further enhances robustness by integrating optical flow for occlusion handling. Alternatively, MOTIP~\cite{gao2024multiple} enables efficient batch parallel training by reformulating tracking as direct ID classification through learnable ID dictionaries. 
However, these methods only implicitly optimize object embeddings via the joint detection and tracking losses, lacking explicit constraints for discriminativeness enhancement, which leads to high inter-object similarity and limited tracking performance.
\section{Methodology}
\label{sec:Methodology}
\subsection{Overview}
As illustrated in \cref{fig:framework}, FDTA builds upon the standard architecture of end-to-end MOT methods~\cite{zeng2022motr,gao2023memotr,zhang2023motrv2,galoaa2024more,gao2024multiple}.
To provide context, we first summarize the process used by end-to-end MOT methods.
Given a sequence of input frames $\{\boldsymbol{I}_t\}_{t=1}^{T}$, a shared DETR module~\cite{carion2020end, zhu2020deformable} is employed to process each frame $t$ independently, generating object embeddings $\{\boldsymbol{e}_i^t\}$ for each object $i$ within the frame.
These embeddings are then passed through an ID prediction module to predict object identities, which are used to associate objects across frames and form object trajectories.

Based on the common patterns of object embeddings observed in \cref{fig:intro}, FDTA explicitly incorporates Spatial Adapter (SA), Temporal Adapter (TA), and Identity Adapter (IA) to enhance the discriminativeness of object embeddings from complementary perspectives.
Each component is detailed in the following sections.

\subsection{Spatial Adapter}
\label{sec:spatial-adaptation}

Robust association requires continuous spatial understanding across frames to distinguish overlapping objects, while depth provides 3D geometric cues to handle occlusion. 
Based on this, Spatial Adapter (SA) distills depth knowledge from large-scale pretrained depth estimators, enriching object embeddings with continuous spatial information.

\begin{figure}
  \centering
  \includegraphics[width=\linewidth]{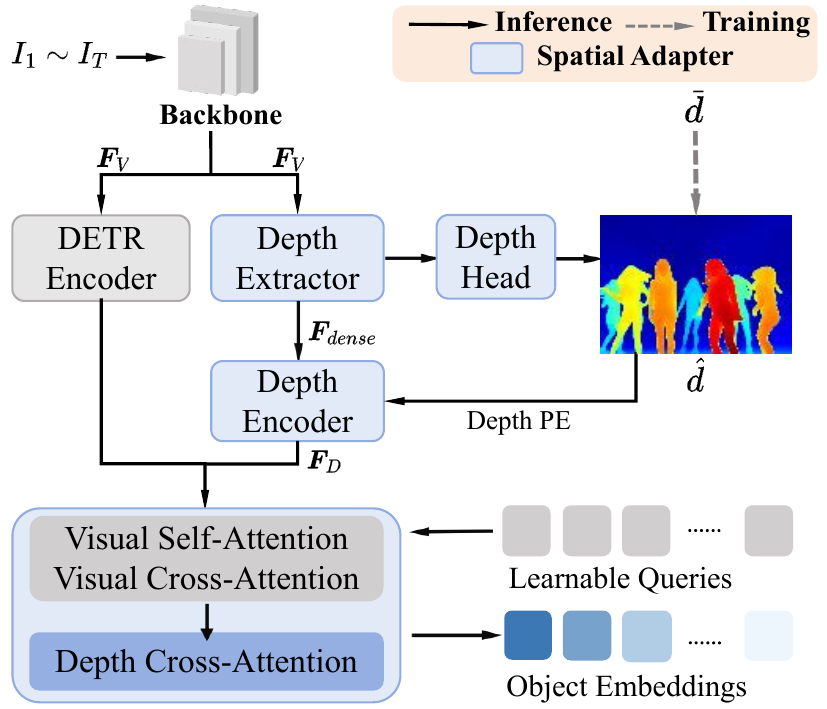}
  \caption{The detailed architecture of Spatial Adapter. }
  \label{fig:spatial-adapter}
\end{figure}

\noindent\textbf{Depth Extraction.}
As illustrated in \cref{fig:spatial-adapter}, parallel to the original DETR, SA employs a two-layer convolutional depth extractor to derive dense features, denoted as $\boldsymbol{{F}}_{dense}$, from the backbone feature $\boldsymbol{{F}}_V$.
Additionally, a single-layer convolutional depth head is employed to predict per-pixel depth probabilities $\boldsymbol{d}$ over discrete bins through Linear-Increasing Discretization (LID)~\cite{zhang2023monodetr}. We then obtain the depth map $\hat{\boldsymbol{d}}$ by weighting the depth bin values with their predicted probabilities. The details are provided in~\cref{subsec:suppl_sa_architecture}.

\noindent\textbf{Depth Distillation.} Benefiting from the generalization capability of pretrained foundation models, we leverage Video Depth Anything~\cite{chen2025video} to generate offline pseudo depth labels. These continuous labels are discretized into bins $\overline{\boldsymbol{d}}$ using LID for supervision. 

Since the tracking task fundamentally prioritizes the object's region over the background, we introduce a weighted depth loss to emphasize the quality of foreground depth. Specifically, we utilize the ground truth bounding box to differentiate between foreground and background pixels, and assign a larger penalty weight $w_{i,j}$ to foreground pixels during training. 
The depth loss is formulated as:
\begin{equation}
\mathcal{L}_{\text{depth}} = \frac{1}{N_{\text{total}}}\sum_{i,j} w_{i,j} \cdot \text{FL}(\boldsymbol{d}_{i,j}, \overline{\boldsymbol{{d}}}_{i,j}),
\label{eq:depth-loss}
\end{equation}
where $\text{FL}(\cdot)$ denotes the Focal Loss~\cite{lin2017focal}, $\boldsymbol{d}_{i,j}$ and $\overline{\boldsymbol{{d}}}_{i,j}$ represent the predicted depth probabilities and discretized ground-truth bin at position $(i,j)$, and $w_{i,j}$ indicates different weights for foreground and background pixels.

\noindent\textbf{Depth Encoding.} 
After obtaining the depth-aware features $\boldsymbol{{F}}_{dense}$ via distillation, we introduce a depth encoding strategy to inject these features into the object embeddings.
Specifically, we first utilize a depth encoder, which mirrors the architecture of the standard DETR encoder, to produce refined depth feature $\boldsymbol{{F}}_D$. Unlike visual features, the predicted depth map $\hat{\boldsymbol{d}}$ provides vital positional information for refinement.
Accordingly, we compute learnable depth positional embeddings $\text{PE}_d$ through linear interpolation:
\begin{equation}
\text{PE}_d = (1-\delta) \cdot \text{PE}[\lfloor\hat{\boldsymbol{d}}\rfloor] + \delta \cdot \text{PE}[\lceil\hat{\boldsymbol{d}}\rceil], \quad \delta = \hat{\boldsymbol{d}} - \lfloor\hat{\boldsymbol{d}}\rfloor,
\end{equation}
where PE denotes the learnable positional embeddings, and $\lceil\cdot\rceil$, $\lfloor\cdot\rfloor$ denote ceiling and floor operations.
In the DETR decoder, we additionally add a depth cross-attention layer following standard visual attention layers. It allows the object queries to attend directly to the refined depth features $\boldsymbol{{F}}_D$. This yields depth-enriched object embeddings with enhanced spatial discrimination.

\begin{figure}[t]
    \centering
    \includegraphics[width=\columnwidth]{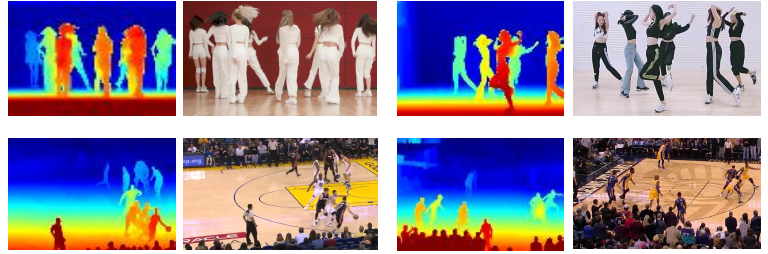}
    \caption{Visualization of predicted depth maps on DanceTrack and SportsMOT.}
    \label{fig:depth}
\end{figure}

\noindent\textbf{Inference Efficiency.} 
Through depth distillation, SA module learns to provide essential depth-aware features $\boldsymbol{{F}}_{D}$, allowing us to deprecate the large foundation model during inference.
Notably, this approach represents the first implementation of object embedding enhancement within an end-to-end tracking paradigm. This design ensures both a low computational cost and accurate depth estimation, effectively addressing the limitations of relying on external depth estimators~\cite{wu2024depthmot,zhao2025detrack} or inaccurate heuristic assumptions based on perspective projections~\cite{pdsort2025,limanta2024camot}.
The detailed computational analysis and depth prediction visualization are provided in \cref{sec:computational_analysis} and \cref{fig:depth}, respectively.

\subsection{Temporal Adapter}
\label{sec:temporal-adapter}

While the SA module effectively enhances the spatial discriminativeness of object embeddings within individual frames, these frame-independent embeddings inherently lack temporal context across the sequence for association.
As a complement, we propose Temporal Adapter (TA) that explicitly captures temporal dependencies across entire trajectories through sequence modeling, enriching embeddings with temporal discriminativeness.

\noindent\textbf{Trajectory Modeling.} For online tracking at frame $t$, we aggregate the historical trajectory $\boldsymbol{{F}}_i^{\text{traj}} = \{\boldsymbol{e}_i^{t-T}, \ldots, \boldsymbol{e}_i^{t-1}\}$ for each identity $i$ across $T$ previous frames. To capture temporal dependencies, we employ a standard transformer encoder~\cite{vaswani2017attention} with $L$ layers to process the sequence. 
However, for standard attention, all tokens attend to each other bidirectionally, which would cause future information leakage and unreliable interactions with [empty] tokens of missing objects.
Based on these characteristics of tracking, we design a specialized dual attention mask $\boldsymbol{M} \in \mathbb{B}^{T\times T}$ that is a union of causal and missing constraints in the binary domain:
\begin{equation}
\boldsymbol{M}[j,k] = \begin{cases} 
1 & \text{if } k > j \text{ or not detected} \\
0 & \text{otherwise}
\end{cases},
\end{equation}
where $j, k$ are frame indices.
Note that diagonal elements are kept unmasked for numerical stability.
The TA module processes the historical trajectory features $\boldsymbol{{F}}_i^{\text{traj}}$ using the custom attention mask $\boldsymbol{M}$ as:
\begin{equation}
    \hat{\boldsymbol{{F}}}_i^{\text{traj}} = \text{TA}(\boldsymbol{{F}}_i^{\text{traj}}, \boldsymbol{M}),
\end{equation}
where $\hat{\boldsymbol{{F}}}_i^{\text{traj}}$ represents the trajectory embeddings that encode temporal dependencies. Visualization of the temporal interaction weight is provided in~\cref{subsec:suppl_additional_results_vis}.
\subsection{Identity Adapter}
\label{sec:identity-adapter}
Up to this point, SA and TA enhance object embeddings from spatial and temporal perspectives within the scope of tracking objectives. 
We further focus on exploring explicit optimization objectives directly aligned with the association task. 
At the instance level, Identity Adapter (IA) leverages quality-aware contrastive learning on the object embeddings to learn discriminative features for identification.
Specifically, IA aims to pull together embeddings of the same object across different frames while pushing apart embeddings of different objects.

\noindent\textbf{Contrastive Pair Sampling.}
We construct positive and negative pairs as the foundation for contrastive learning. 
Specifically, we first aggregate all object embeddings across frames to form the full sample pool. 
Each object embedding $\boldsymbol{e}_i^s$ at frame $s$ is assigned an identity label $\mathrm{ID}_i^s$ using Hungarian matching~\cite{kuhn1955hungarian} with the ground-truth bounding boxes. 
We define positive pair set $\mathcal{P}$ as embeddings that share the same identity, and negative pair set $\mathcal{N}$ as those with different identities.
Each pair of embeddings $(\boldsymbol{e}_i^s, \boldsymbol{e}_j^k)$ is categorized as:
\begin{equation}
\begin{aligned}
\mathcal{P} &= \big\{(\boldsymbol{e}_i^s, \boldsymbol{e}_j^k) \ \big| \ k \neq s, \ 
\mathrm{ID}_i^s = \mathrm{ID}_j^k \big\}, \\
\mathcal{N} &= \big\{(\boldsymbol{e}_i^s, \boldsymbol{e}_j^k) \ \big| \ \mathrm{ID}_i^s \neq \mathrm{ID}_j^k \big\}.
\end{aligned}
\end{equation}
This sampling strategy allows object embeddings to interact with all others within and across frames. 
Analogous to the benefits of enlarged sample pools observed in MoCo~\cite{he2020momentum}, 
our design substantially increases the number of negative samples for tracking scenarios.

\noindent\textbf{Quality-Aware Contrastive Learning.} 
Building upon the sampled positive and negative pairs, we apply contrastive learning to effectively enhance object embeddings through two modular designs.

\noindent
\textit{IoU-Filter.}
In tracking tasks, the samples are generated from predictions, where not all embeddings are equally reliable. 
Hence, quality control is essential. 
During identity assignment in sample curation, each object embedding computes its Intersection-over-Union (IoU) score with the corresponding ground truth, which serves as a quality metric.
Specifically, we retain only high-quality embeddings with $\text{IoU}_i^t \geq 0.5$ to filter out noisy samples.
For each positive pair $(\boldsymbol{e}_i^s, \boldsymbol{e}_j^k) \in \mathcal{P}$, we further assign a weight using the harmonic mean of their IoU scores:
\begin{equation}
w(\boldsymbol{e}_i^s,\boldsymbol{e}_j^k) = \frac{2 \cdot \text{IoU}_i^s \cdot \text{IoU}_j^k}{\text{IoU}_i^s + \text{IoU}_j^k}.    
\end{equation}

\noindent
\textit{Consistent Feature Extraction.}
Since object embeddings contain frame-variant cues such as motion and pose, directly applying contrastive learning on these embeddings would interfere with leveraging temporal information.
To address this, we employ a 3-layer MLP $\phi$ to extract identity-consistent features from these embeddings.

Based on the above components, the final quality-aware contrastive learning loss is defined as:
\begin{equation}
\mathcal{L}_{\text{IA}} = \frac{1}{|\mathcal{P}|} \sum_{(\boldsymbol{e}_i^s,\boldsymbol{e}_j^k) \in \mathcal{P}} w(\boldsymbol{e}_i^s,\boldsymbol{e}_j^k) \cdot \mathcal{L}_{\text{InfoNCE}}(\boldsymbol{e}_i^s,\boldsymbol{e}_j^k).
\label{eq:IA-loss}
\end{equation}
Here, $\mathcal{L}_{\text{InfoNCE}}$ denotes the standard InfoNCE loss~\cite{oord2018cpc}:
\begin{equation}
\mathcal{L}_{\text{InfoNCE}}(\boldsymbol{e}_i^s,\boldsymbol{e}_j^k) = -\log \frac{\exp(\phi(\boldsymbol{e}_i^s) \cdot \phi(\boldsymbol{e}_j^k)/\tau)}{\sum_{\boldsymbol{e}\in \mathcal{E}}\exp(\phi(\boldsymbol{e}_i^s) \cdot \phi(\boldsymbol{e})/\tau)},
\end{equation}
where the set $\mathcal{E}=\left\{\boldsymbol{e}| (\boldsymbol{e}_{i}^{s},\boldsymbol{e}) \in \mathcal{N}\right\} \cup \left\{\boldsymbol{e}_{j}^k\right\}$ contains the positive sample $\boldsymbol{e}_{j}^k$ and all negative samples of $\boldsymbol{e}_{i}^{s}$. $\tau$ is the temperature parameter. Note that contrastive learning is applied only during training, thus introducing no additional inference overhead. 
The IA module is the first to perform quality-aware contrastive learning in an end-to-end tracking framework, clearly distinguishing it from tracking-by-detection methods~\cite{fischer2023qdtrack,yu2022discriminative,somers2025cameltrack}.

\subsection{Loss Function}
\label{sec:loss}

FDTA is trained end-to-end with a combined loss:
\begin{equation}
    \mathcal{L} = \mathcal{L}_{\text{det}} + \lambda_{\text{ID}}\mathcal{L}_{\text{ID}} + \lambda_{\text{depth}}\mathcal{L}_{\text{depth}} + \lambda_{\text{IA}}\mathcal{L}_{\text{IA}},
\end{equation}
where $\mathcal{L}_{\text{det}}$ comprises standard DETR losses~\cite{zhu2020deformable}. 
$\mathcal{L}_{\text{ID}}$ is the cross-entropy loss for ID classification. $\mathcal{L}_{\text{depth}}$ and
$\mathcal{L}_{\text{IA}}$ correspond to the depth loss \cref{eq:depth-loss} and quality-aware contrastive loss \cref{eq:IA-loss}, respectively. Loss coefficients $\lambda$ balance each component.

\section{Experiment}
\label{sec:Experiment}
\subsection{Experimental Setup}
\begin{table}[t]
 \centering
 \setlength{\tabcolsep}{2pt}
 \caption{
 Performance comparison with state-of-the-art methods on DanceTrack test set. 
 The best result for each metric is shown in \textbf{bold}. $^*$ indicates methods using extra training data. 
 }
 \footnotesize{
 \begin{tabular}{l@{\hspace{2pt}}ccc|cc}
     \toprule
     \textbf{Methods} & \textbf{HOTA}$\uparrow$ & \textbf{IDF1}$\uparrow$ & \textbf{AssA}$\uparrow$ & \textbf{MOTA}$\uparrow$ & \textbf{DetA}$\uparrow$ \\
     \midrule
    \textit{\underline{Tracking-by-Detection:}} \\
    SORT~\cite{bewley2016simple} (ICIP2016) & 47.9 & 50.8 & 31.2 & 91.8 & 72.0 \\
    DeepSORT~\cite{wojke2017simple} (ICIP2017) & 45.6 & 47.9 & 29.7 & 87.8 & 71.0 \\
    CenterTrack~\cite{zhou2020tracking} (ECCV2020) & 41.8 & 35.7 & 22.6 & 86.8 & 78.1 \\
    FairMOT~\cite{zhang2021fairmot} (IJCV2021) & 39.7 & 40.8 & 23.8 & 82.2 & 66.7 \\
    QDTrack~\cite{fischer2023qdtrack} (CVPR2021) & 45.7 & 44.8 & 29.2 & 83.0 & 72.1 \\
    ByteTrack~\cite{zhang2022bytetrack} (ECCV2022) & 47.3 & 52.5 & 31.4 & 89.5 & 71.6 \\
    OC-SORT~\cite{cao2022observation} (CVPR2023) & 55.1 & 54.2 & 38.0 & 89.4 & 80.3 \\
    Hybrid-SORT~\cite{yang2024hybrid} (AAAI2024) & 65.7 & 67.4 & --- & 91.8 & --- \\
    SparseTrack~\cite{liu2025sparsetrack} (TCSVT2025) & 55.7 & 58.1 & 39.3 & 91.3 & 79.2 \\
    DiffMOT~\cite{lv2024diffmot} (CVPR2025) & 62.3 & 63.0 & 47.2 & 92.8 & 82.5 \\
    TrackTrack~\cite{shim2025focusing} (CVPR2025) & 66.5 & 67.8 & 52.9 & \textbf{93.6} & --- \\
     \midrule
    \textit{\underline{End-to-End:}} \\
    TransTrack~\cite{sun2020transtrack} (arXiv2020) & 45.5 & 45.2 & 27.5 & 88.4 & 75.9 \\
    MOTR~\cite{zeng2022motr} (ECCV2022) & 54.2 & 51.5 & 40.2 & 79.7 & 73.5 \\
    MeMOTR~\cite{gao2023memotr} (ICCV2023) & 63.4 & 65.5 & 52.3 & 85.4 & 77.0 \\
    CO-MOT~\cite{yan2023bridging} (CVPR2023) & 69.4 & 71.9 & 58.9 & 91.2 & 82.1 \\
    MOTRV2~\cite{zhang2023motrv2} (CVPR2023) & 69.9 & 71.7 & 59.0 & 91.9 & \textbf{83.0} \\
    MOTIP~\cite{gao2024multiple} (CVPR2025) & 67.5 & 72.2 & 57.6 & 90.3 & 79.4 \\
    SambaMOTR~\cite{segu2024samba} (ICLR2025) & 67.2 & 70.5 & 57.5 & 88.1 & 78.8 \\
    \rowcolor{gray!20} Ours & \textbf{71.7} & \textbf{77.2} & \textbf{63.5} & 91.3 & 81.0 \\
     \midrule
    \textit{\underline{with extra data:}} \\
    MOTRV2$^*$~\cite{zhang2023motrv2} (CVPR2023) & 73.4 & 76.0 & 64.4 & 92.1 & 83.7 \\
    MOTRv3$^*$~\cite{li2023motrv3} (arXiv2023) & 70.4 & 72.3 & 59.3 & \textbf{92.9} & \textbf{83.8} \\
    MOTIP$^*$~\cite{gao2024multiple} (CVPR2025) & 71.4 & 76.3 & 62.8 & 91.6 & 81.3 \\
    \rowcolor{gray!20} Ours$^*$ & \textbf{74.4} & \textbf{80.0} & \textbf{67.0} & 92.2 & 82.7 \\
     \bottomrule
 \end{tabular}
 }
 \label{tab:dancetrack-comparison}
 \vspace{-3mm}
\end{table}

\noindent\textbf{Datasets.} 
We evaluate FDTA on three challenging benchmarks with similar appearance and complex movements. DanceTrack~\cite{sun2020dancetrack} contains 100 dance videos where dancers wear identical clothing and perform complex synchronized movements. SportsMOT~\cite{cui2023sportsmot} consists of 240 sequences from basketball, volleyball, and soccer scenes, featuring fast-paced motion and frequent occlusions in competitive sports. BFT~\cite{zheng2024nettrack} includes 106 bird flock clips from the BBC documentary Earthflight, featuring complex aerial dynamics, varying formations, and large-scale flocking behavior.

\noindent\textbf{Evaluation Metrics.} The evaluation follows standard MOT protocols, using comprehensive metrics including Higher Order Tracking Accuracy (HOTA)~\cite{luiten2021hota}, ID F1 Score (IDF1)~\cite{ristani2016performance}, Association Accuracy (AssA), Multi-Object Tracking Accuracy (MOTA)~\cite{bernardin2008evaluating}, and Detection Accuracy (DetA). Among these, HOTA, IDF1, and AssA primarily assess tracking performance.

\subsection{Implementation Details}
FDTA is built upon Deformable DETR~\cite{zhu2020deformable} with ResNet-50~\cite{he2016deep} backbone, initialized with COCO~\cite{lin2014microsoft} pretrained weights. We train with video sequences of $T=30$ frames per batch. For SA, the foreground weighting factor $w$ is set to 7. For TA, the encoder consists of $L=6$ layers. For IA, we set the temperature parameter $\tau=0.1$. We train FDTA for 11 epochs with batch size 4 on 4 NVIDIA H200 GPUs. 
Following prior work~\cite{segu2024samba,zeng2022motr,zhang2023motrv2,gao2024multiple}, we apply standard data augmentation, including random resize, crop, and flip, along with trajectory occlusion and identity switching. We use AdamW optimizer~\cite{loshchilov2017decoupled} with initial learning rate $1 \times 10^{-4}$ and weight decay $5 \times 10^{-4}$. The loss weights are set as 
$\lambda_{\text{depth}}=1.0$, $\lambda_{\text{ID}}=1.0$, and $\lambda_{\text{IA}}=1.0$ based on validation performance. 

\begin{table}[t]
 \centering
 \setlength{\tabcolsep}{2pt}
 \caption{
 Performance comparison with state-of-the-art methods on SportsMOT test set. 
 The best result for each metric is shown in \textbf{bold}.
 }
 \footnotesize{
 \begin{tabular}{l@{\hspace{2pt}}ccc|cc}
     \toprule
     \textbf{Methods} & \textbf{HOTA}$\uparrow$ & \textbf{IDF1}$\uparrow$ & \textbf{AssA}$\uparrow$ & \textbf{MOTA}$\uparrow$ & \textbf{DetA}$\uparrow$ \\
     \midrule
    \textit{\underline{Tracking-by-Detection:}} \\
    CenterTrack~\cite{zhou2020tracking} (ECCV2020) & 62.7 & 60.0 & 48.0 & 90.8 & 82.1 \\
    FairMOT~\cite{zhang2021fairmot} (IJCV2021) & 49.3 & 53.5 & 34.7 & 86.4 & 70.2 \\
    QDTrack~\cite{fischer2023qdtrack} (CVPR2021) & 60.4 & 62.3 & 47.2 & 90.1 & 77.5 \\
    GTR~\cite{zhou2022global} (CVPR2022) & 54.5 & 55.8 & 45.9 & 67.9 & 64.8 \\
    ByteTrack~\cite{zhang2022bytetrack} (ECCV2022) & 62.8 & 69.8 & 51.2 & 94.1 & 77.1 \\
    BoT-SORT~\cite{aharon2022bot} (arXiv2022) & 68.7 & 70.0 & 55.9 & \textbf{94.5} & 84.4 \\
    OC-SORT~\cite{cao2022observation} (CVPR2023) & 68.1 & 68.0 & 54.8 & 93.4 & 84.8 \\
    DiffMOT~\cite{lv2024diffmot} (CVPR2025) & 72.1 & 72.8 & 60.5 & \textbf{94.5} & \textbf{86.0} \\
     \midrule
    \textit{\underline{End-to-End:}} \\
    TransTrack~\cite{sun2020transtrack} (arXiv2020) & 68.9 & 71.5 & 57.5 & 92.6 & 82.7 \\
    MeMOTR~\cite{gao2023memotr} (ICCV2023) & 68.8 & 69.9 & 57.8 & 90.2 & 82.0 \\
    MOTIP~\cite{gao2024multiple} (CVPR2025) & 71.9 & 75.0 & 62.0 & 92.9 & 83.4 \\
    SambaMOTR~\cite{segu2024samba} (ICLR2025) & 69.8 & 71.9 & 59.4 & 90.3 & 82.2 \\
    \rowcolor{gray!20} Ours & \textbf{74.2} & \textbf{78.5} & \textbf{65.5} & 93.0 & 84.1 \\
     \bottomrule
 \end{tabular}
 }
 \label{tab:sportsmot-comparison}
\end{table}
\begin{figure*}[t]
  \centering
  \includegraphics[width=\textwidth]{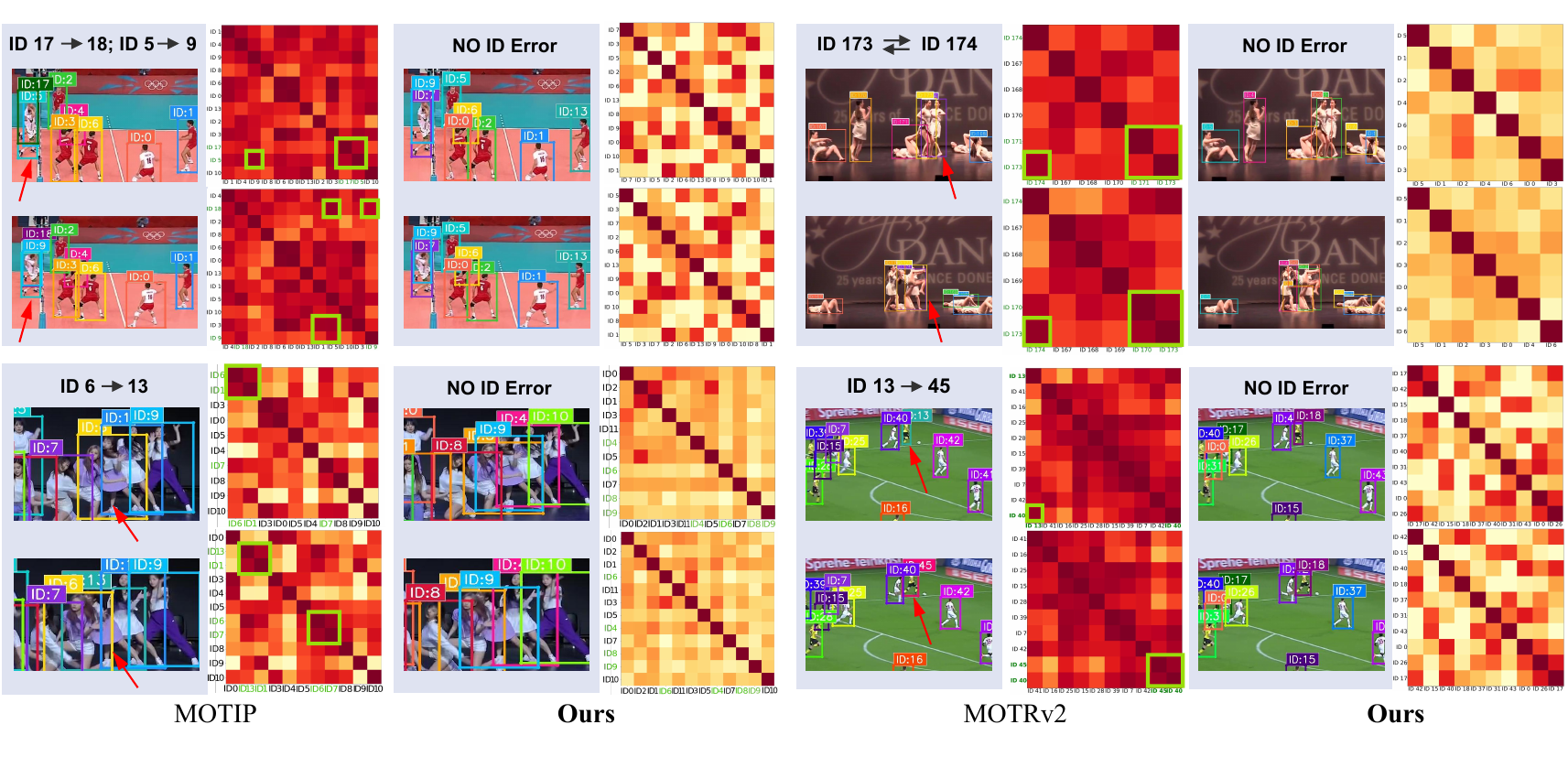}
  \vspace{-6mm}
  \caption{Tracking results visualization and inter-object embedding similarity matrix comparison on DanceTrack and SportsMOT. Darker colors in the similarity matrix indicate higher similarity. Green boxes in the similarity matrix highlight high-similarity regions. Incorrect IDs are marked in the tracking results.}
  \label{fig:tracking_vis}
\end{figure*}

\begin{figure*}[t]
  \centering
  \includegraphics[width=\linewidth]{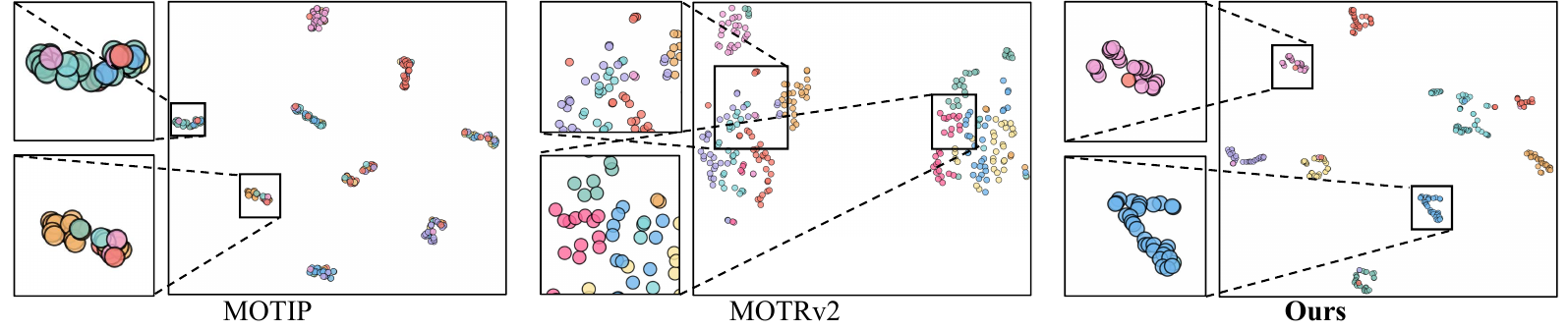}
  \vspace{-6mm}
  \caption{t-SNE visualization of object embeddings on DanceTrack sequence dancetrack0015. Each color represents a tracked object.}
  \label{fig:tsne}
  \vspace{-2mm}
\end{figure*}

\subsection{State-of-the-Art Comparison}
\noindent\textbf{DanceTrack.} As shown in \cref{tab:dancetrack-comparison}, FDTA achieves state-of-the-art performance with 71.7\% HOTA, 77.2\% IDF1, and 63.5\% AssA, outperforming the previous best method MOTRv2 by +1.8\%, +5.5\%, and +4.5\% respectively. Training with additional validation data further improves performance to 74.4\% HOTA and 80.0\% IDF1 on the test set. These gains in association metrics are notable, considering DanceTrack's extreme inter-object similarity from identical clothing and synchronized movements, which highlights the effectiveness of our approach. Qualitative visual comparisons are provided in \cref{fig:tracking_vis}.

\noindent\textbf{SportsMOT.} As shown in \cref{tab:sportsmot-comparison}, FDTA achieves state-of-the-art performance. The improvements on association metrics demonstrate the effectiveness of FDTA in handling fast-paced motion and frequent occlusions in competitive sports.

\begin{table}[t]
 \centering
 \setlength{\tabcolsep}{2pt}
 \caption{
 Performance comparison with state-of-the-art methods on BFT test set. 
 The best result for each metric is shown in \textbf{bold}.
 }
 \footnotesize{
 \begin{tabular}{l@{\hspace{2pt}}ccc|cc}
     \toprule
     \textbf{Methods} & \textbf{HOTA}$\uparrow$ & \textbf{IDF1}$\uparrow$ & \textbf{AssA}$\uparrow$ & \textbf{MOTA}$\uparrow$ & \textbf{DetA}$\uparrow$ \\
     \midrule
    \textit{\underline{Tracking-by-Detection:}} \\
    SORT~\cite{bewley2016simple} (ICIP2016) & 61.2 & 77.2 & 62.3 & 75.5 & 60.6 \\
    JDE~\cite{wang2020towards} (ECCV2020) & 30.7 & 37.4 & 23.4 & 35.4 & 40.9 \\
    CenterTrack~\cite{zhou2020tracking} (ECCV2020) & 65.0 & 61.0 & 54.0 & 60.2 & 58.5 \\
    FairMOT~\cite{zhang2021fairmot} (IJCV2021) & 40.2 & 41.8 & 28.2 & 56.0 & 53.3 \\
    CSTrack~\cite{liang2022CSTrack} (TIP2022) & 33.2 & 34.5 & 23.7 & 46.7 & 47.0 \\
    ByteTrack~\cite{zhang2022bytetrack} (ECCV2022) & 62.5 & 82.3 & 64.1 & 77.2 & 61.2 \\
    OC-SORT~\cite{cao2022observation} (CVPR2023) & 66.8 & 79.3 & 68.7 & 77.1 & 65.4 \\
     \midrule
     \textit{\underline{End-to-End:}} \\
     TransCenter~\cite{xu2021transcenter} (arXiv2021) & 60.0 & 72.4 & 61.1 & 74.1 & 66.0 \\
     TransTrack~\cite{sun2020transtrack} (arXiv2020) & 62.1 & 71.4 & 60.3 & 71.4 & 64.2 \\
     TrackFormer~\cite{meinhardt2022trackformer} (CVPR2022) & 63.3 & 72.4 & 61.1 & 74.1 & 66.0 \\
     SambaMOTR~\cite{segu2024samba} (ICLR2025) & 69.6 & 81.9 & 73.6 & 72.0 & 66.0 \\
     MOTIP~\cite{gao2024multiple} (CVPR2025) & 70.5 & 82.1 & 71.8 & 77.1 & 69.6 \\
     \rowcolor{gray!20} Ours & \textbf{72.2} & \textbf{84.2} & \textbf{74.5} & \textbf{78.2} & \textbf{70.1} \\
     \bottomrule
 \end{tabular}
 }
 \label{tab:bft-comparison}
 \vspace{-5mm}
\end{table}
\noindent\textbf{BFT.} To further validate on non-human scenarios, we evaluate on BFT featuring large flocks with extreme density and rapid formation changes. As shown in \cref{tab:bft-comparison}, FDTA achieves state-of-the-art performance across all metrics. The consistent improvements across diverse scenarios, from human crowds to competitive sports to bird flocks, demonstrate that our approach is broadly applicable to a wide range of tracking scenarios.

\subsection{Qualitative Analysis}

\noindent\textbf{Tracking Results and Embedding Visualization.} 
We visualize tracking results and inter-object embedding similarity matrix of two strong baselines, MOTIP~\cite{gao2024multiple} and MOTRv2~\cite{zhang2023motrv2}, on DanceTrack and SportsMOT. As shown in \cref{fig:tracking_vis}, both baseline methods exhibit ID errors (red arrows). By examining their embedding similarity matrix, we find these ID errors consistently correlate with high inter-object similarity in the embedding space (green boxes). This validates our key insight that high feature similarity between different objects degrades tracking performance. In contrast, our method produces more discriminative embeddings with reduced inter-object similarity, achieving stable tracking with fewer ID errors. More visualizations are provided in ~\cref{subsec:suppl_additional_results_vis}.

\noindent\textbf{Embedding Space Visualization.} 
To further analyze embedding discriminability, we apply t-SNE to visualize object embeddings across a tracking sequence, where each color represents one tracked object. Unlike single-frame similarity matrix, t-SNE shows temporal embedding patterns across multiple frames. As shown in \cref{fig:tsne}, existing methods exhibit two critical issues: (1) overlapping clusters where different objects lack clear boundaries, and (2) mixed clusters containing embeddings from multiple objects, while our method groups same-identity embeddings tightly together and separates different identities distinctly.

\subsection{Ablation Studies}
\label{sec:ablation}

We conduct ablation studies on the DanceTrack test set to validate our approach, including module-level ablations and detailed design choices within each adapter.

\noindent\textbf{Complementary Effects of Three Adapters.} To evaluate each adapter's contribution, we progressively add them as shown in \cref{tab:main_ablation_study}. Rows 2-4 show that each adapter individually improves performance. Rows 5-7 show that combining any two adapters yields further improvements. Our complete FDTA (Row 8) achieves 71.7\% HOTA and 77.2\% IDF1, outperforming all other combinations and showing that enhancing object embeddings from spatial, temporal, and identity perspectives is effective and complementary.

\noindent\textbf{Spatial Adapter Design.} 
We examine the effectiveness of Depth Positional Encoding by enabling and disabling this module. As shown in \cref{tab:depth_fusion_ablation}, when directly using raw depth features without Depth PE, performance drops by 0.4\% HOTA and 0.3\% IDF1, demonstrating that encoding depth as learnable positional embeddings further enhances spatial understanding. Additional ablation studies on foreground weighting and depth encoding strategies are provided in~\cref{subsec:suppl_spatial_ablation}.

\noindent\textbf{Temporal Adapter Design.}
TA models temporal dependencies through trajectories with a causal mask to prevent information leakage. We examine the impact of handling missing objects in Table~\ref{tab:trajectory_timing_ablation}. Replacing missing objects with zero vectors (Row 2) degrades performance by 1.3\% HOTA and 2.9\% IDF1, performing worse than without TA (Row 1). Our missing mask that distinguishes present and missing objects (Row 3) improves performance by 1.0\% HOTA and 1.2\% IDF1, demonstrating that proper handling of missing objects is crucial for trajectory modeling. Further ablations demonstrating that TA leverages temporal information are provided in ~\cref{subsec:suppl_temporal_ablation}.

\begin{table}[t]
  \centering
  \caption{Ablation study on the proposed adapters. SA: Spatial Adapter; TA: Temporal Adapter; IA: Identity Adapter.}
  \vspace{-2mm}
\resizebox{\columnwidth}{!}{
\footnotesize
\renewcommand{\arraystretch}{0.75}
\begin{tabular}{c@{\hspace{8pt}}c@{\hspace{4pt}}c@{\hspace{4pt}}c|c@{\hspace{4pt}}c@{\hspace{4pt}}c@{\hspace{4pt}}c@{\hspace{4pt}}c}
\toprule
 & \textbf{SA} & \textbf{TA} & \textbf{IA} & \textbf{HOTA}$\uparrow$ & \textbf{IDF1}$\uparrow$ & \textbf{AssA}$\uparrow$ & \textbf{MOTA}$\uparrow$ & \textbf{DetA}$\uparrow$ \\
\midrule
1 & \xmark    & \xmark    & \xmark    & 69.4 & 74.5 & 60.2 & 90.6 & 80.0 \\
2 & \cmark & \xmark    & \xmark    & 70.2 & 74.8 & 61.2 & 90.9 & 80.7 \\
3 & \xmark    & \cmark & \xmark    & 70.4 & 75.7 & 61.3 & \textbf{91.3} & 81.1 \\
4 & \xmark    & \xmark    & \cmark & 70.1 & 74.8 & 60.7 & 91.2 & \textbf{81.2} \\
5 & \cmark & \cmark & \xmark    & 70.8 & 76.8 & 61.9 & 91.2 & \textbf{81.2} \\
6 & \cmark & \xmark & \cmark & 70.6 & 75.7 & 62.0 & 91.1 & 80.8 \\
7 & \xmark & \cmark & \cmark & 71.0 & 76.5 & 62.2 & \textbf{91.3} & \textbf{81.2} \\
\rowcolor{gray!20}
8 & \cmark & \cmark & \cmark & \textbf{71.7} & \textbf{77.2} & \textbf{63.5} & \textbf{91.3} & 81.0 \\
\bottomrule
\end{tabular}
}
\label{tab:main_ablation_study}
\vspace{-2mm}
\end{table}
\begin{table}[t]
  \centering
  \caption{Ablation study on Depth PE in Spatial Adapter.}
  \vspace{-2mm}
\resizebox{\columnwidth}{!}{
\footnotesize
\renewcommand{\arraystretch}{0.2}
\begin{tabular}{l|@{\hspace{8pt}}c@{\hspace{4pt}}c@{\hspace{4pt}}c@{\hspace{4pt}}c@{\hspace{4pt}}c}
\toprule
\textbf{Depth Encoding} & \textbf{HOTA}$\uparrow$ & \textbf{IDF1}$\uparrow$ & \textbf{AssA}$\uparrow$ & \textbf{MOTA}$\uparrow$ & \textbf{DetA}$\uparrow$ \\
\midrule
w/o Depth PE & 71.3 & 76.9 & 63.1 & 91.2 & 81.0 \\
w/ Depth PE & \textbf{71.7} & \textbf{77.2} & \textbf{63.5} & \textbf{91.3} & 81.0 \\
\bottomrule
\end{tabular}
}
\label{tab:depth_fusion_ablation}
\end{table}
\begin{table}[t]
  \centering
    \caption{Ablation study on dual attention mask in Temporal Adapter.}
\resizebox{\columnwidth}{!}{
\normalsize
\renewcommand{\arraystretch}{0.75}
\begin{tabular}{l|@{\hspace{6pt}}c@{\hspace{8pt}}c@{\hspace{8pt}}c@{\hspace{8pt}}c@{\hspace{8pt}}c}
\toprule
\textbf{Missing Object Handling} & \textbf{HOTA}$\uparrow$ & \textbf{IDF1}$\uparrow$ & \textbf{AssA}$\uparrow$ & \textbf{MOTA}$\uparrow$ & \textbf{DetA}$\uparrow$ \\
\midrule
w/o TA & 69.4 & 74.5 & 60.2 & 90.6 & 80.0 \\
w/ Zero Vector & 68.1 & 71.6 & 57.5 & 91.1 & 80.9 \\
w/ Missing Mask & \textbf{70.4} & \textbf{75.7} & \textbf{61.3} & \textbf{91.3} & \textbf{81.1} \\
\bottomrule
\end{tabular}
}
\label{tab:trajectory_timing_ablation}
\vspace{-2mm}
\end{table}

\noindent\textbf{Identity Adapter Design.} We evaluate three essential components of IA in \cref{tab:contrastive_ablation}: Contrastive Learning (CL), Consistent Feature Extractor (CFE), and IoU-Filter (IF). Directly applying CL on raw embeddings (Row 2) degrades performance by 1.0\% HOTA and 1.6\% IDF1 compared to Row 1. This occurs because raw embeddings contain frame-variant cues that interfere with learning stable identity features. Introducing CFE (Row 3) projects embeddings into an identity-specific space, improving performance by 0.5\% HOTA over Row 1 and recovering the degradation from Row 2 by 1.5\% HOTA and 1.6\% IDF1. This demonstrates that extracting stable identity features is essential for effective contrastive learning. Adding IF (Row 4) further improves performance by 0.2\% HOTA and 0.3\% IDF1 over Row 3, confirming that IoU-based sample weighting effectively enhances the contrastive objective by prioritizing reliable positive pairs.

\subsection{Computational Analysis}
\label{sec:computational_analysis}
To assess the computational overhead introduced by our proposed components, Table~\ref{tab:inference_breakdown} provides a detailed breakdown of inference time on the DanceTrack at 1920$\times$1080 resolution using an H200 GPU. SA and TA (highlighted in gray) introduce minimal overhead of only 1.4\% and 2.7\%, respectively. Notably, IA is absent from this table as it functions solely during training, incurring zero inference cost. DETR dominates at 83.9\%, inherent to the base architecture. Overall,  FDTA maintains comparable speed at 13.4 FPS (74.76 ms per frame), improving tracking performance without compromising speed.

\begin{table}[t]
  \centering
  \caption{Ablation study on Identity Adapter. CL: Contrastive Learning; CFE: Consistent Feature Extractor; IF: IoU-Filter.}
\resizebox{\columnwidth}{!}{
\footnotesize
\renewcommand{\arraystretch}{0.6}
\begin{tabular}{c@{\hspace{6pt}}c@{\hspace{6pt}}c|@{\hspace{8pt}}c@{\hspace{8pt}}c@{\hspace{8pt}}c@{\hspace{8pt}}c@{\hspace{8pt}}c}
\toprule
\textbf{CL} & \textbf{CFE} & \textbf{IF} & \textbf{HOTA}$\uparrow$ & \textbf{IDF1}$\uparrow$ & \textbf{AssA}$\uparrow$ & \textbf{MOTA}$\uparrow$ & \textbf{DetA}$\uparrow$ \\
\midrule
\xmark & \xmark & \xmark & 69.4 & 74.5 & 60.2 & 90.6 & 80.0 \\
\cmark & \xmark & \xmark & 68.4 & 72.9 & 58.3 & 90.7 & 80.6 \\
\cmark & \cmark & \xmark & 69.9 & 74.5 & 60.5 & 91.0 & 80.8 \\
\rowcolor{gray!20}\cmark & \cmark & \cmark & \textbf{70.1} & \textbf{74.8} & \textbf{60.7} & \textbf{91.2} & \textbf{81.2} \\
\bottomrule
\end{tabular}
}
\label{tab:contrastive_ablation}
\end{table}
\begin{table}[t]
  \centering
  \caption{Computational breakdown on 1920$\times$1080 resolution. Gray rows indicate our proposed adapters.}
\footnotesize
\renewcommand{\arraystretch}{0.50}
\setlength{\tabcolsep}{8pt}
\begin{tabular}{lcc}
\toprule
\textbf{Component} & \textbf{Time (ms)} & \textbf{\% of Total} \\
\midrule
DETR & 62.71 & 83.9\% \\
\rowcolor{gray!25}SA  & 1.02 & 1.4\% \\
\rowcolor{gray!25}TA  & 1.99 & 2.7\% \\
ID Prediction & 6.67 & 8.9\% \\
Other Components & 2.37 & 3.2\% \\
\midrule
Total & 74.76 & 100\% \\
\bottomrule
\end{tabular}
\label{tab:inference_breakdown}
\vspace{-2mm}
\end{table}

\section{Conclusion}
We identify that object embeddings generated by shared DETR exhibit severe inter-object similarity, which inadequately serves tracking requirements. To address this, we propose FDTA to explicitly enhance object embeddings through three complementary perspectives: Spatial Adapter for 3D geometric understanding, Temporal Adapter for temporal modeling, and Identity Adapter for instance-level separation. Extensive experiments demonstrate that FDTA achieves state-of-the-art performance across DanceTrack, SportsMOT, and BFT, while maintaining efficient inference speed. Future work will leverage foundation models such as video generation and world models to synthesize challenging corner cases for enhancing tracking robustness in extreme scenarios.

\section*{Acknowledgments}
The work is supported by Shanghai Artificial Intelligence Laboratory.

{
    \small
    \bibliographystyle{ieeenat_fullname}
    \bibliography{main}
}

\clearpage
\setcounter{page}{1}
\maketitlesupplementary

\renewcommand{\thesection}{\Alph{section}}
\setcounter{section}{0}

\section{Overview}
\label{sec:suppl_overview}

In the supplementary material, we primarily:
\begin{enumerate}
    \item Present design rationale and implementation details in \cref{sec:suppl_architecture}.
    \item Present experimental details including dataset processing, training strategies, and loss function in \cref{sec:suppl_exp_details}.
    \item Provide additional experimental analysis and visualization results in \cref{sec:suppl_analysis}.
\end{enumerate}

\section{FDTA Details}
\label{sec:suppl_architecture}

In this section, we present the design rationale and implementation details for the Spatial Adapter (\cref{subsec:suppl_sa_architecture}), Temporal Adapter (\cref{subsec:suppl_ta_architecture}), and Identity Adapter (\cref{subsec:suppl_ia_architecture}).

\subsection{Spatial Adapter Design}
\label{subsec:suppl_sa_architecture}

\subsubsection{Depth Extractor}
\label{subsubsec:depth_extractor}

The depth extractor fuses pyramid features from the backbone $\boldsymbol{F}_V$. Similar to the visual branch, we use three feature levels with spatial strides of 8, 16, and 32 (denoted as $\mathbf{f}_8$, $\mathbf{f}_{16}$, and $\mathbf{f}_{32}$) to capture both fine-grained spatial details and global scene context for depth estimation.

To derive the dense features $\boldsymbol{F}_{dense}$, we first project features from each scale to a unified dimension $c=256$ using $1\times1$ convolutional layers. The coarser-scale features at strides of 16 and 32 are then upsampled to the finest resolution via bilinear interpolation and averaged:
\begin{equation}
\begin{split}
    \boldsymbol{{F}}_{\text{avg}} &= \frac{1}{3}(\mathbf{f}_8 + \mathbf{f}_{16}^{\uparrow} + \mathbf{f}_{32}^{\uparrow})
\end{split}
\end{equation}

The averaged features $\boldsymbol{{F}}_{\text{avg}}$ are then processed through two $3\times3$ convolutional blocks to produce the final dense features $\boldsymbol{{F}}_{dense}$ for depth prediction.

\subsubsection{Depth Discretization}
\label{subsubsec:depth_discretization}

Inspired by MonoDETR~\cite{zhang2023monodetr}, we formulate depth prediction as a classification task for stable training. This requires discretizing the continuous pseudo depth values from Video Depth Anything~\cite{chen2025video} into depth bins, each representing a depth range. We employ Linear-Increasing Discretization (LID) to partition these bins. 

We compute the bin size as:
\begin{equation}
    \text{bin\_size} = \frac{2(d_{\max} - d_{\min})}{K(1 + K)}
\end{equation}
For the $K$ foreground bins, the depth values are computed as:
\begin{equation}
    b_i = (i + 0.5)^2 \cdot \frac{\text{bin\_size}}{2} - \frac{\text{bin\_size}}{8} + d_{\min}
\end{equation}
where $i = 0, 1, \ldots, K-1$.
For the last bin corresponding to background, we set $b_K = d_{\max}$. Here $d_{\min} = 10^{-3}$ and $d_{\max} = 256$ define the depth range. Unlike uniform binning, LID allocates more bins to near depth ranges where tracking is more critical.

\subsubsection{Depth Prediction Head}
\label{subsubsec:depth_head}

For depth prediction, the depth head predicts a probability distribution over the $K+1$ bins for each pixel, which is then converted to continuous depth values. Specifically, a single-layer $1\times1$ convolutional head maps the dense features $\boldsymbol{{F}}_{dense}$ to $(K+1)$ channels. After softmax, we obtain the per-pixel probability distribution ${\boldsymbol{d}}$. The final continuous depth map $\hat{\boldsymbol{d}}$ is computed via weighted summation:
\begin{equation}
    \hat{\boldsymbol{d}} = \sum_{i=0}^{K} d_i \cdot b_i
\end{equation}
where $d_i$ is the predicted probability for the $i$-th bin and $b_i$ represents the depth value of that bin.  
\begin{figure}[t]
    \centering
    \includegraphics[width=\linewidth]{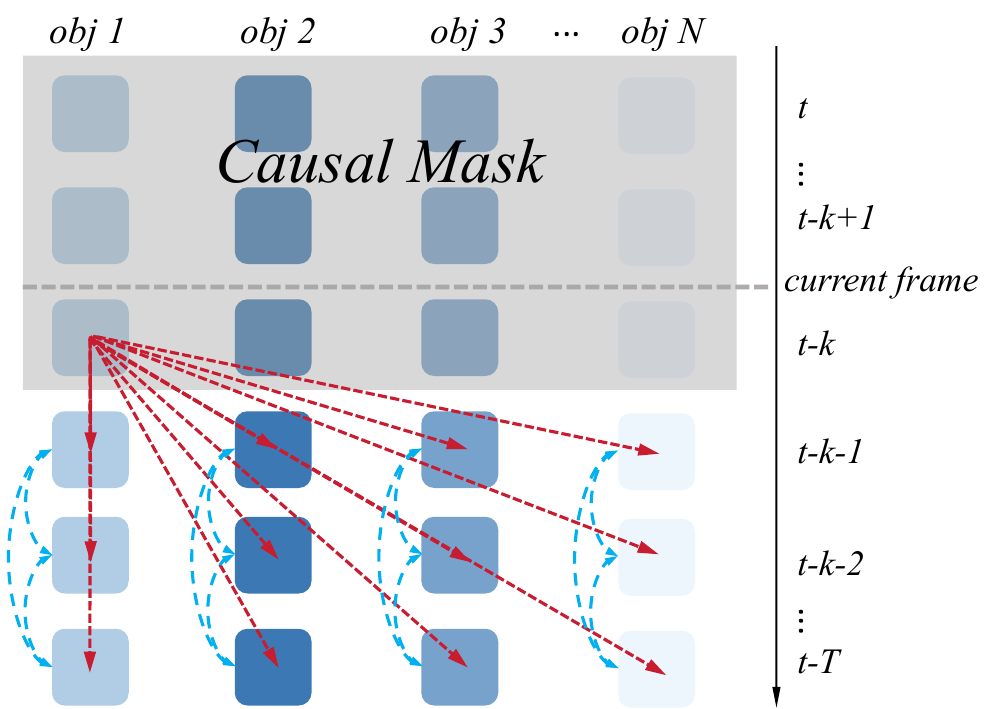}
    \caption{Object embedding interaction in ID prediction and TA. Red arrows represent the interaction in ID prediction, where objects at the current frame query historical embeddings for identity matching. Blue arrows represent the interaction in TA, where embeddings within the same trajectory interact across frames to enrich temporal context.}
    \label{fig:embedding_interaction}
\end{figure}
\subsection{Temporal Adapter Design Motivation}
\label{subsec:suppl_ta_architecture}

\label{subsubsec:embedding_interaction}
\begin{figure*}[t]
    \centering
    \includegraphics[width=\linewidth]{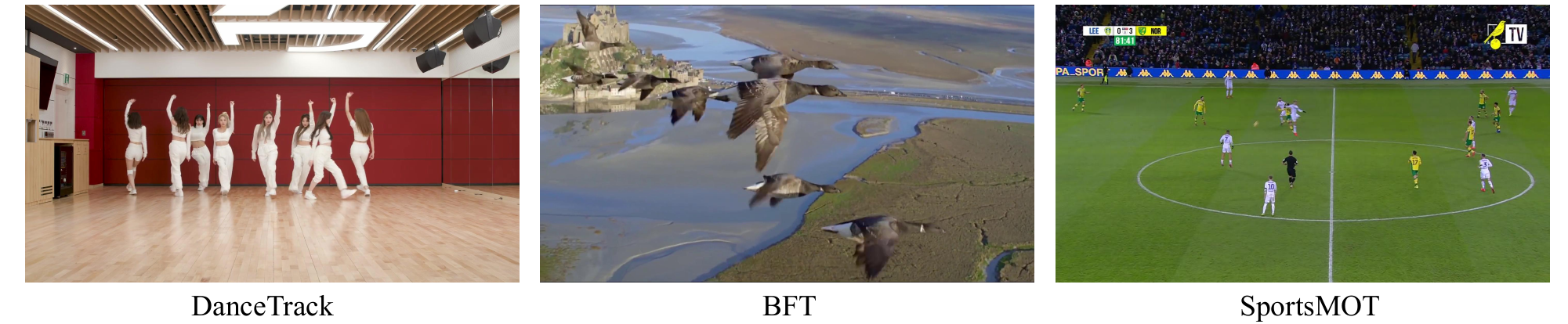}
    \caption{Visual examples from different datasets used in our experiments.}
    \label{fig:suppl_datasets}
\end{figure*}
As illustrated in Figure~\ref{fig:embedding_interaction}, the ID prediction assigns identities via cross-attention where objects at current frame $t-k$ query all historical object embeddings from preceding frames $[t-k-1, \ldots, t-T]$ to retrieve the closest match (red arrows). However, each historical embedding is independently encoded without inter-frame interaction, containing only information from its own frame. To address this limitation, we propose the Temporal Adapter (TA) to enable interaction among historical embeddings before ID prediction. Through temporal modeling across historical frames (blue arrows), each object embedding aggregates information from its trajectory history, thereby enriching embeddings with temporal context for more discriminative identity matching.

\subsection{Identity Adapter Design}
\label{subsec:suppl_ia_architecture}

MoCo~\cite{he2020momentum} demonstrates that more positive pairs improve contrastive learning. However, previous methods~\cite{fischer2023qdtrack,somers2025cameltrack} typically perform contrastive learning only between consecutive frames, yielding limited positive pairs per embedding. We expand positive pair selection by leveraging all embeddings across the entire training batch from multiple trajectories and frames. Given an identity appearing in $M$ frames, any two from different frames form a positive pair, yielding $M(M-1)/2$ pairs. This significantly increases training samples and provides richer supervision for learning discriminative features.

\begin{figure*}[!ht]
    \centering
    \includegraphics[width=\linewidth]{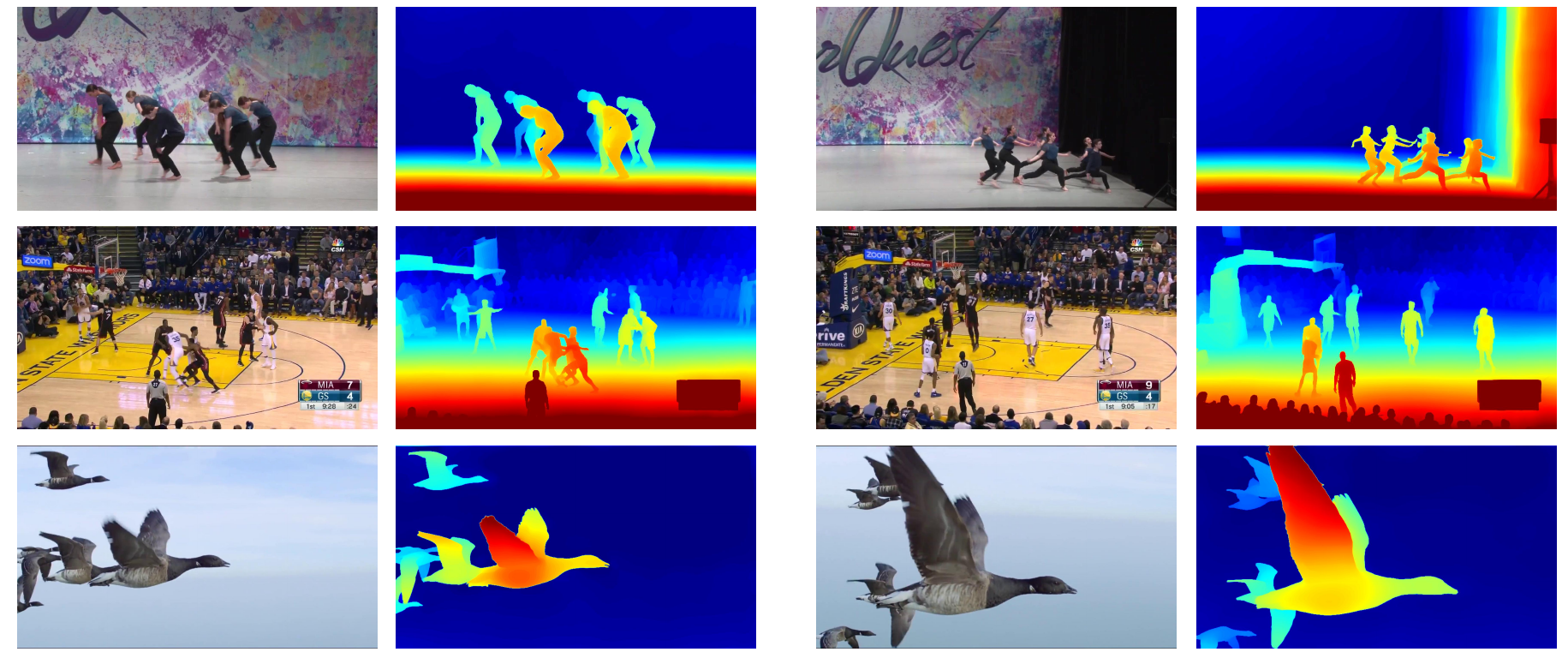}
    \caption{Pseudo depth labels visualization on DanceTrack, SportsMOT, and BFT.}
    \label{fig:suppl_depth_quality}
\end{figure*}

\section{Experimental Details}
\label{sec:suppl_exp_details}

This section provides detailed experimental configurations that complement the main text, including dataset processing (\cref{subsec:suppl_dataset_processing}), training implementation details (\cref{subsec:suppl_training_strategies}), and loss function details (\cref{subsec:suppl_loss_details}).

\subsection{Dataset Processing}
\label{subsec:suppl_dataset_processing}

We evaluate our method on three diverse benchmarks: DanceTrack, SportsMOT, and BFT, strictly following standard protocols~\cite{zeng2022motr,meinhardt2022trackformer,gao2023memotr,zhang2023motrv2,galoaa2024more,gao2024multiple} with official train/val/test splits. Specifically, DanceTrack consists of 40 training sequences, 25 validation sequences, and 35 test sequences; SportsMOT contains 45 training sequences, 45 validation sequences, and 150 test sequences; BFT provides 45 training sequences, 25 validation sequences, and 36 test sequences. Visual examples from these datasets are provided in \cref{fig:suppl_datasets}.

For training, following standard protocols~\cite{zeng2022motr,zhang2023motrv2,gao2023memotr,gao2024multiple}, we adopt a sequence-based sampling strategy where each training sample consists of a trajectory with length $T=30$. To enable robust learning of motion dynamics across varying speeds, we sample frames with random temporal intervals between 1 and 4, rather than using fixed consecutive frames. The same trajectory length $T$ is maintained during inference to ensure consistency. 

For input data, we use original images with a resolution of $1920\times1080$, and generate corresponding depth maps at the same resolution offline using Video Depth Anything~\cite{chen2025video} with the ViT-L backbone. During training, consistent data augmentation is applied to both RGB images and depth maps to guarantee strict spatial alignment.

\subsection{Training Details}
\label{subsec:suppl_training_strategies}
We adopt several key strategies for efficient and stable training. Specifically, we inject sinusoidal position encodings~\cite{vaswani2017attention} into the Temporal Adapter to capture temporal order. Additionally, to mitigate the instability of DETR-based detectors in early training stages, we adopt a warm-up strategy for the Identity Adapter. We train for 11 epochs in total, and disable the contrastive loss in the first epoch to ensure learning from reliable object features.

\subsection{DETR Loss Function}
\label{subsec:suppl_loss_details}

Following standard practices in Deformable DETR~\cite{zhu2020deformable}, the detection loss combines three components:
\begin{equation}
\mathcal{L}_{\text{det}} = \lambda_{\text{cls}}\mathcal{L}_{\text{cls}} + \lambda_{\text{bbox}}\mathcal{L}_{\text{bbox}} + \lambda_{\text{giou}}\mathcal{L}_{\text{giou}},
\end{equation}
Bounding box regression consists of an $L_1$ loss measuring coordinate differences and a GIoU loss capturing spatial overlap:
\begin{equation}
    \mathcal{L}_{\text{bbox}} = \|b_{\text{pred}} - b_{\text{gt}}\|_1, \quad
    \mathcal{L}_{\text{giou}} = 1 - \text{GIoU}(b_{\text{pred}}, b_{\text{gt}}).
\end{equation}
Classification is optimized using focal loss:
\begin{equation}
\mathcal{L}_{\text{cls}} = -\alpha_t(1-p_t)^\gamma \log(p_t),
\end{equation}
where $p_t$ is the estimated probability for the target class, $\alpha_t$ weights different classes, and $\gamma$ controls the focus on hard examples. We keep the same hyperparameters as Deformable DETR~\cite{zhu2020deformable}: $\lambda_{\text{cls}} = 2.0$, $\lambda_{\text{bbox}} = 5.0$, $\lambda_{\text{giou}} = 2.0$, $\alpha_t = 0.25$, and $\gamma = 2$.

\section{Additional Experimental Analysis}
\label{sec:suppl_analysis}

\subsection{Pseudo Depth Label Generation Analysis}
\label{subsec:suppl_depth_generation}

\subsubsection{Computational Cost and Storage Requirements}
\label{subsubsec:depth_cost_storage}

We generate pseudo depth labels offline using Video Depth Anything~\cite{chen2025video}, eliminating online inference overhead. For the entire training set, the inference takes approximately 1 hour on an H200 GPU for DanceTrack, 0.6 hours for SportsMOT, and 0.4 hours for BFT. The storage requirements are reasonable, with approximately 8.4 GB for DanceTrack, 3.4 GB for SportsMOT, and 1.4 GB for BFT. This one-time cost is negligible compared to the training time, and the generated depth labels can be reused across multiple experiments.

\subsubsection{Pseudo Depth Quality}
\label{subsubsec:depth_quality}

As shown in \cref{fig:suppl_depth_quality}, we visualize the pseudo depth labels generated by Video Depth Anything~\cite{chen2025video} across three diverse datasets. The generated depth maps exhibit high quality, maintaining consistent depth estimation both within individual frames and across temporal sequences. Additionally, they effectively distinguish objects at varying spatial positions. This provides reliable supervision for tracking and discriminative cues for embedding enhancement.

\subsection{Discussion on Performances of Tracking-by-Detection Methods}

In comparison with SOTA methods (see \cref{tab:dancetrack-comparison,tab:sportsmot-comparison,tab:bft-comparison}), tracking-by-detection methods~\cite{aharon2022bot,shim2025focusing,lv2024diffmot} achieve high MOTA and DetA scores compared to end-to-end methods~\cite{zeng2022motr,gao2023memotr,zhang2023motrv2,gao2024multiple}, but show relatively lower performance in metrics such as AssA, IDF1, and HOTA. This is because different metrics emphasize different aspects of tracking capability.

Understanding what each metric measures is key to proper evaluation. MOTA heavily weighs detection errors, while DetA directly evaluates detection accuracy. Tracking-by-detection methods naturally excel in these detection-focused metrics by leveraging powerful standalone detectors like YOLOX~\cite{ge2021yolox}. However, the core challenge in tracking is maintaining consistent identities across frames, not just detection. Metrics such as HOTA~\cite{luiten2021hota}, which equally weights detection and association accuracy, and association-focused metrics (AssA, IDF1) that specifically evaluate the ability to maintain consistent identities, better reflect tracking capability. Our method achieves significant improvements on these association metrics, demonstrating strong tracking performance.

\subsection{Additional Ablations on Spatial Adapter}
\label{subsec:suppl_spatial_ablation}

\subsubsection{Depth Encoding Layer Design}
\label{subsubsec:depth_layer_design}

To integrate depth information into object embeddings, we investigate the optimal placement of the depth cross-attention layer within the decoder. As shown in \cref{tab:depth_ablation}, we insert the depth cross-attention layer (\textit{Depth}) at different positions in each decoder block of the standard DETR decoder, which contains self-attention (\textit{Self}) and visual cross-attention (\textit{Vision}). We compare three placements: before self-attention (Row 2), between self-attention and visual cross-attention (Row 3), and after visual cross-attention (Row 4), with Row 1 showing the performance without depth. Results show that placing depth cross-attention after visual cross-attention (\textit{Self} $\rightarrow$ \textit{Vision} $\rightarrow$ \textit{Depth}) achieves the best performance with 0.8\% HOTA improvement. Therefore, we adopt this configuration in our Spatial Adapter.
\subsubsection{Foreground Weighting Factor}

\begin{table}[t]
  \centering
  \caption{Ablation study on depth encoding layer designs in SA. \textit{Self}, \textit{Vision}, and \textit{Depth} denote self-attention, visual cross-attention, and depth cross-attention layers.}
\footnotesize{
\begin{tabular}{>{\itshape}l|c@{\hspace{3pt}}c@{\hspace{3pt}}c@{\hspace{3pt}}c@{\hspace{3pt}}c}
\toprule
\textbf{Architecture} & \textbf{HOTA}$\uparrow$ & \textbf{IDF1}$\uparrow$ & \textbf{AssA}$\uparrow$ & \textbf{MOTA}$\uparrow$ & \textbf{DetA}$\uparrow$ \\
\midrule
Self $\rightarrow$ Vision & 69.4 & 74.5 & 60.2 & 90.6 & 80.0 \\
Depth $\rightarrow$ Self $\rightarrow$ Vision & 68.2 & 72.4 & 58.1 & 90.4 & 80.2 \\
Self $\rightarrow$ Depth $\rightarrow$ Vision & 69.3 & 73.7 & 59.6 & \textbf{91.2} & 80.7 \\
Self $\rightarrow$ Vision $\rightarrow$ Depth & \textbf{70.2} & \textbf{74.8} & \textbf{61.2} & 90.9 & \textbf{80.7} \\
\bottomrule
\end{tabular}
}
\label{tab:depth_ablation}
\end{table}

\begin{table}[t]
  \centering
  \caption{Ablation study on foreground weighting in SA.}
  \footnotesize{
\begin{tabular}{l|c@{\hspace{3pt}}c@{\hspace{3pt}}c@{\hspace{3pt}}c@{\hspace{3pt}}c}
\toprule
\textbf{Setting} & \textbf{HOTA}$\uparrow$ & \textbf{IDF1}$\uparrow$ & \textbf{AssA}$\uparrow$ & \textbf{MOTA}$\uparrow$ & \textbf{DetA}$\uparrow$ \\
\midrule
w/o Foreground Weighting & 70.7 & 75.7 & 61.6 & \textbf{91.4} & \textbf{81.3} \\
w/ Foreground Weighting & \textbf{71.7} & \textbf{77.2} & \textbf{63.5} & 91.3 & 81.0 \\
\bottomrule
\end{tabular}
}
\label{tab:foreground_weight}
\end{table}

As shown in \cref{tab:foreground_weight}, we evaluate foreground weighting in depth loss, which assigns larger weights to pixels within object bounding boxes during training. This strategy encourages the model to focus on learning accurate depth for foreground objects, which is more critical for tracking. The results show 1.0\% HOTA and 1.5\% IDF1 improvement, demonstrating its effectiveness.

\subsection{Temporal Adapter Design Ablation}
\label{subsec:suppl_temporal_ablation}

To verify that TA effectively leverages temporal information, we evaluate different trajectory history lengths as shown in \cref{tab:trajectory_length_ablation}, where the blue superscripts denote the performance gains from using TA. We observe that TA brings larger improvements with longer history. Specifically, at 5 frames, TA improves HOTA by +0.7\%, while at 30 frames, the gain increases to +1.0\% HOTA and +1.2\% IDF1. This trend demonstrates TA's practical effectiveness. Longer sequences contain richer temporal information, and TA's attention mechanism enables each object embedding to aggregate information across all historical frames, enriching embeddings with comprehensive temporal context for better association. We adopt 30 frames as the default setting, balancing performance and training efficiency.

\begin{table}[t]
  \centering
  \caption{Ablation study on trajectory history length in TA. The blue superscripts denote the performance gains from using TA.}
\resizebox{\columnwidth}{!}{
\footnotesize
\renewcommand{\arraystretch}{0.85}
\begin{tabular}{c@{\hspace{6pt}}c@{\hspace{4pt}}c@{\hspace{8pt}}c@{\hspace{6pt}}c@{\hspace{6pt}}c}
\toprule
\multirow{2}{*}{Length} & \multicolumn{2}{c}{w/o TA} & \multicolumn{2}{c}{w/ TA} & \multirow{2}{*}{\makecell{Time\\(h/ep)}} \\
\cmidrule(lr){2-3} \cmidrule(lr){4-5}
 & HOTA$\uparrow$ & IDF1$\uparrow$ & HOTA$\uparrow$ & IDF1$\uparrow$ & \\
\midrule
5 & 62.7 & 62.8 & 63.4\textsuperscript{\textcolor{blue}{\tiny +0.7}} & 62.8\textsuperscript{\textcolor{blue}{\tiny +0.0}} & 1.0 \\
10 & 66.6 & 67.9 & 67.2\textsuperscript{\textcolor{blue}{\tiny +0.6}} & 69.2\textsuperscript{\textcolor{blue}{\tiny +1.3}} & 1.4 \\
20 & 68.7 & 72.7 & 69.6\textsuperscript{\textcolor{blue}{\tiny +0.9}} & 73.7\textsuperscript{\textcolor{blue}{\tiny +1.0}} & 2.0 \\
\rowcolor{gray!20}
30 & 69.4 & 74.5 & \textbf{70.4}\textsuperscript{\textcolor{blue}{\tiny +1.0}} & \textbf{75.7}\textsuperscript{\textcolor{blue}{\tiny +1.2}} & 2.9 \\
\bottomrule
\end{tabular}
}
\label{tab:trajectory_length_ablation}
\end{table}

\subsection{Additional Experimental Results}
\label{subsec:suppl_additional_results_vis}

\subsubsection{Depth Attention Visualization}
\label{subsubsec:depth_attention_vis}

\begin{figure*}[!ht]
    \centering
    \includegraphics[width=\linewidth]{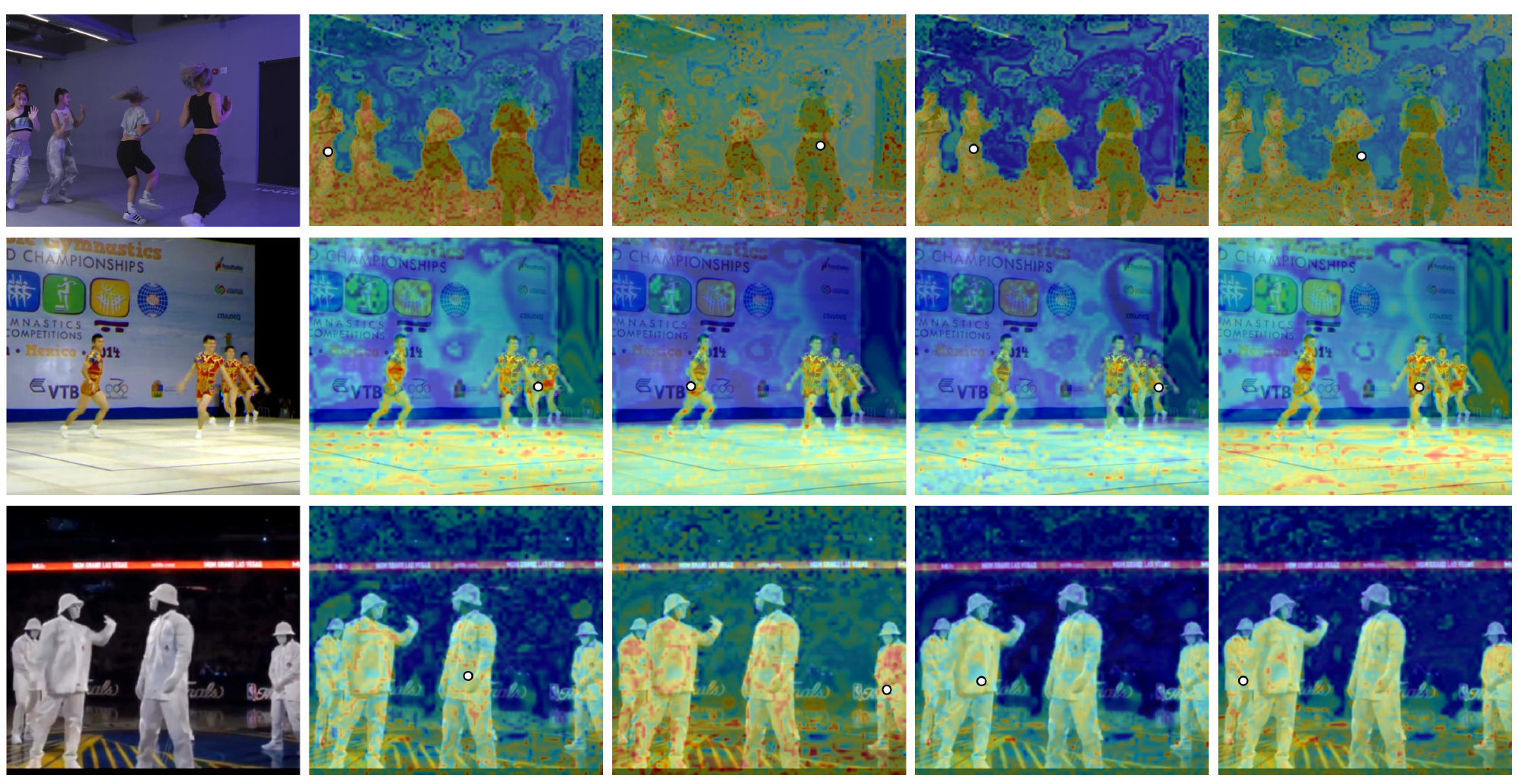}
    \caption{Visualizations of attention maps in SA. The first column denotes the input image, and the last four columns denote the attention maps of the target queries (denoted as white dots). Warmer colors indicate higher attention weights.}
    \label{fig:attention_vis}
  \end{figure*}

To understand how SA leverages depth information, we visualize the attention maps of the depth cross-attention layer in the last decoder block, as shown in \cref{fig:attention_vis}. For each target query (marked as white dots), the attention maps concentrate on foreground objects at similar depth values while suppressing background regions. By aggregating such depth-aware features, object queries obtain enriched embeddings with enhanced discriminativeness for better association.

\subsubsection{Interaction Weight Visualization}
\label{subsubsec:interaction_weight_vis}

\begin{figure*}[!ht]
    \centering
    \includegraphics[width=\linewidth]{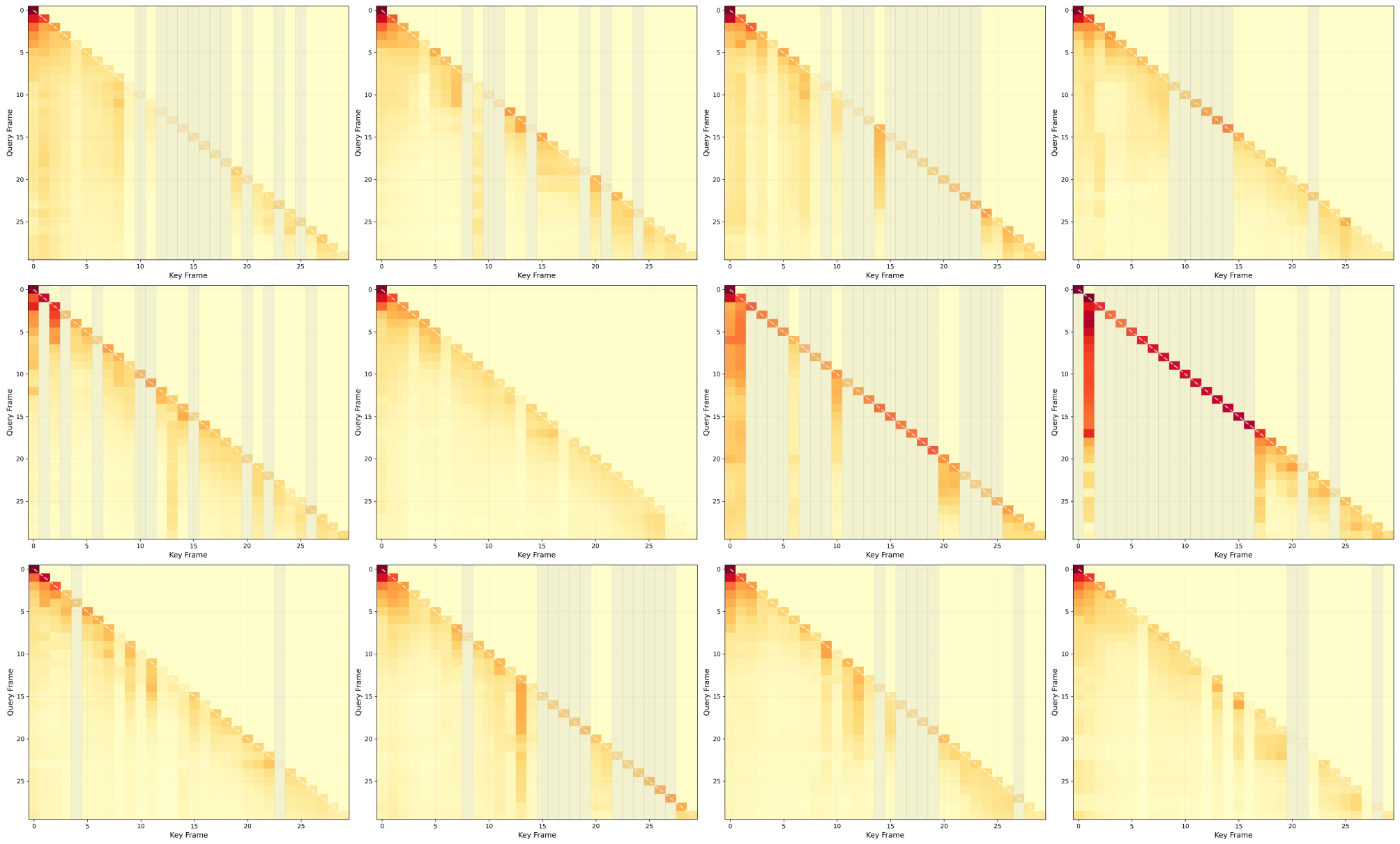}
    \caption{Visualizations of attention weight matrix in TA. Rows represent query frames (current frame) and columns represent key frames (historical frames). Warmer colors indicate higher attention weights. Blue vertical stripes indicate frames where the object is absent, and the lower triangular structure results from the causal mask.}
    \label{fig:ta_weight}
  \end{figure*}

We visualize the temporal interaction weight matrix in TA, as shown in \cref{fig:ta_weight}. Each matrix shows how a tracked object queries its historical frames, where rows represent the query frame (current frame) and columns represent key frames (historical frames). 
The visualization result confirms that the proposed dual-mask strategy achieves its intended effect:
The lower triangular structure results from the causal constraint, ensuring each frame only attends to past frames. Blue vertical stripes mark frames where the object is absent and correctly masked, showing that the mechanism handles undetected objects. Notably, the attention weights are not concentrated on adjacent frames but distributed across the full trajectory, indicating that TA effectively captures long-range temporal dependencies for comprehensive trajectory modeling.

\begin{figure*}[!ht]
  \centering
  \includegraphics[width=0.85\linewidth]{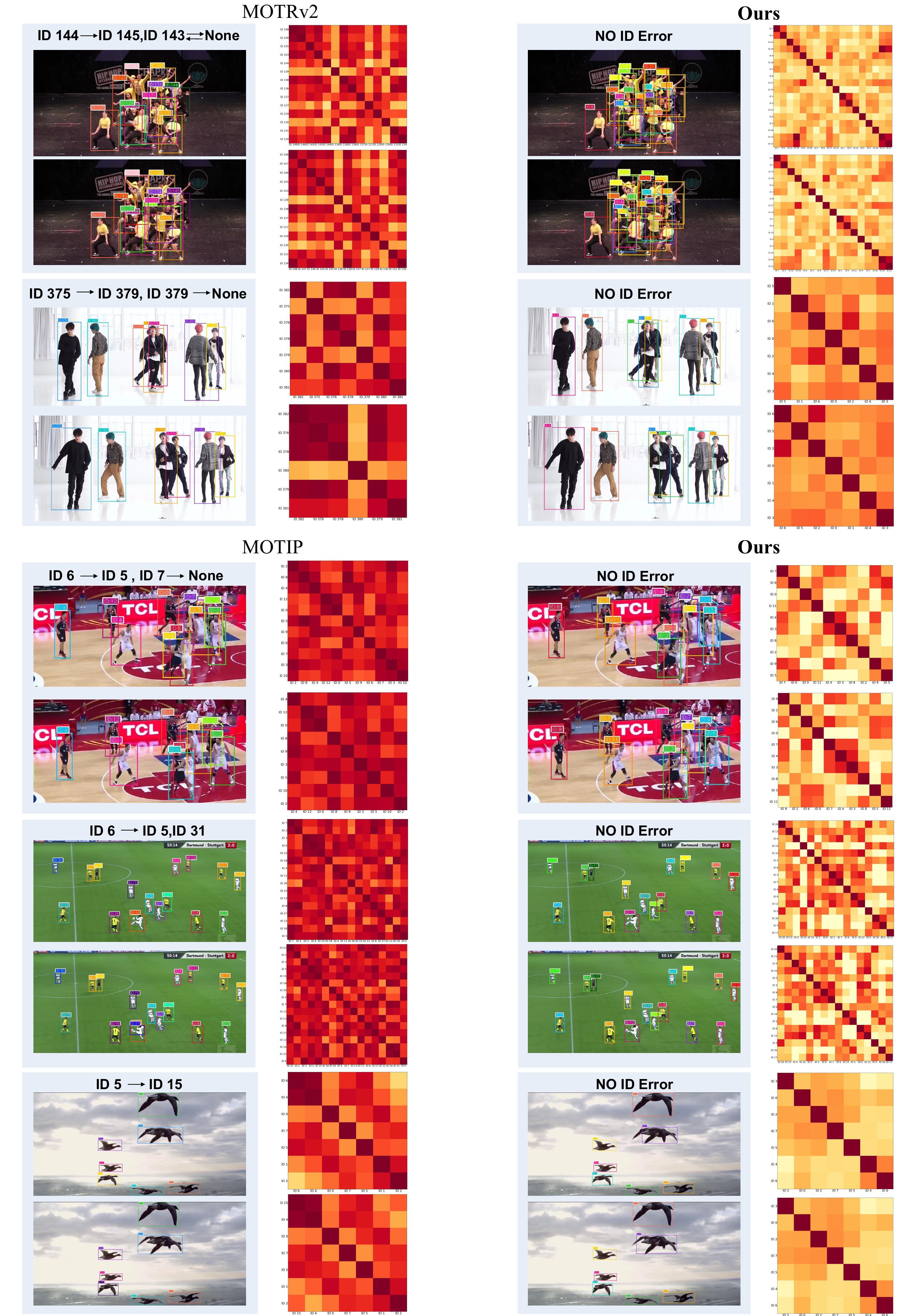}
  \caption{More tracking results visualization and inter-object embedding similarity matrix comparison on DanceTrack, SportsMOT, and BFT. Darker colors in the similarity matrix indicate higher similarity.}
  \label{fig:similarity_vis}
\end{figure*}

\subsubsection{More Tracking Results and Embedding Visualization}
\label{subsubsec:more_results}

We provide additional tracking results and inter-object embedding similarity matrix visualizations on DanceTrack, SportsMOT, and BFT, as shown in \cref{fig:similarity_vis}. The visualizations reveal that baseline methods, MOTRv2 and MOTIP, exhibit high inter-object similarity in the embedding space, leading to tracking errors. In contrast, our method produces more discriminative embeddings by reducing inter-object similarity, achieving more stable tracking across diverse scenarios.

\end{document}